\documentclass[journal]{IEEEtran}
\usepackage{times}
\usepackage{epsfig}
\usepackage{graphicx}
\usepackage{multirow}
\usepackage{amsmath}
\usepackage{amssymb}
\usepackage{algorithm}
\usepackage{algorithmic}
\usepackage{threeparttable}
\usepackage{dsfont}
\usepackage{subfigure}
\usepackage{booktabs}
\usepackage{amsmath,amsthm,amsfonts,amssymb,bm,color}
\ifCLASSINFOpdf
\else
\fi
\begin{document}
%
\title{They are Not Completely Useless: Towards Recycling Transferable Unlabeled Data for Class-Mismatched Semi-Supervised Learning}
%
%
%

\author{Zhuo~Huang,
        Ying~Tai,
        Chengjie~Wang,
        Jian~Yang,~\IEEEmembership{Member,~IEEE}
        Chen~Gong,~\IEEEmembership{Member,~IEEE}
\thanks{Z. Huang, J. Yang, and C. Gong (corresponding author) are with the PCA lab, Key Laboratory of Intelligent Perception and Systems for High-Dimensional Information of Ministry of Education, Nanjing University of Science and Technology, Nanjing 210094, China. (e-mail: hzhuo@njust.edu.cn; csjyang@njust.edu.cn; chen.gong@njust.edu.cn)}
\thanks{Y. Tai and C. Wang are with the Youtu Lab, Tencent. (e-mail: yingtai@tencent.com; jasoncjwang@tencent.com)}
}

%
%

\markboth{Journal of \LaTeX\ Class Files,~Vol.~xx, No.~xx, August~2021}%
{Shell \MakeLowercase{\textit{et al.}}: Bare Demo of IEEEtran.cls for IEEE Journals}
%



\maketitle

\begin{abstract}
Semi-Supervised Learning (SSL) with mismatched classes deals with the problem that the classes-of-interests in the limited labeled data are only a subset of the classes in massive unlabeled data. As a result, classical SSL methods would be misled by the classes which are only possessed by the unlabeled data. To solve this problem, some recent methods divide unlabeled data to useful in-distribution (ID) data and harmful out-of-distribution (OOD) data, among which the latter should particularly be weakened. As a result, the potential value contained by OOD data is largely overlooked. To remedy this defect, this paper proposes a ``Transferable OOD data Recycling'' (TOOR) method which properly utilizes ID data as well as the ``recyclable'' OOD data to enrich the information for conducting class-mismatched SSL. Specifically, TOOR treats the OOD data that have a close relationship with ID data and labeled data as recyclable, and employs adversarial domain adaptation to project them to the space of ID data and labeled data. In other words, the recyclability of an OOD datum is evaluated by its transferability, and the recyclable OOD data are transferred so that they are compatible with the distribution of known classes-of-interests. Consequently, our TOOR extracts more information from unlabeled data than existing methods, so it achieves an improved performance which is demonstrated by the experiments on typical benchmark datasets.
\end{abstract}


%
\IEEEpeerreviewmaketitle

\vspace{-4mm}
\section{Introduction}
%
%
%
%
\IEEEPARstart{T}{he} shortage of labeled data has become a noticeable bottleneck for training many machine learning or computer vision models, as manually annotating a large number of data points is usually prohibitive due to the unaffordable monetary cost or huge demand in human resources. A popular way to deal with such a problem is Semi-Supervised Learning (SSL)~\cite{chapelle2009semi}, which effectively harnesses scarce labeled data and abundant unlabeled data to train an accurate classifier.

\begin{figure}[t]
	\begin{center}
		\includegraphics[width=0.9\linewidth]{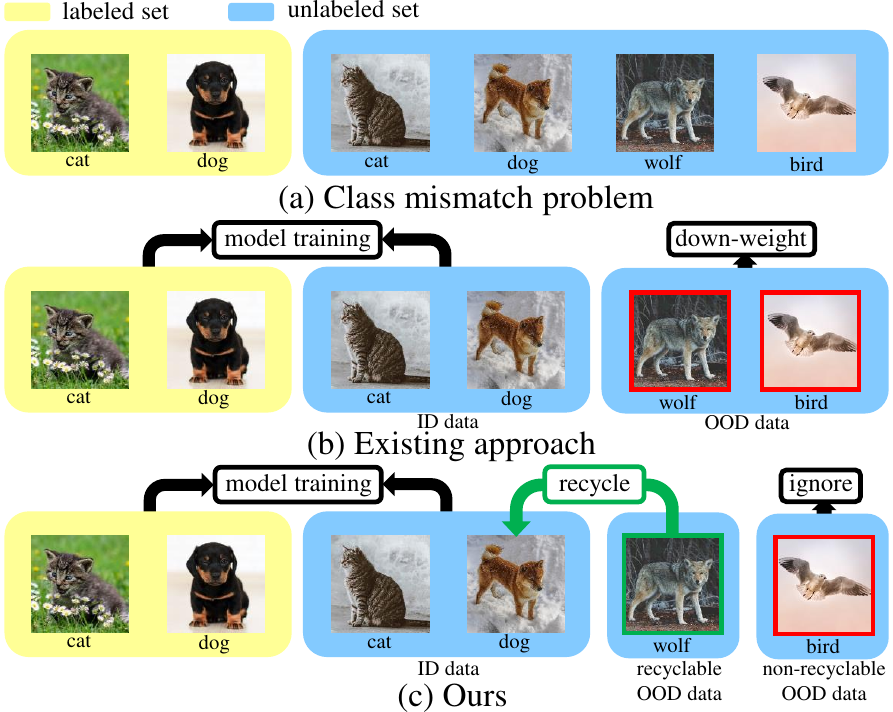}
	\end{center}
\vspace{-3mm}
	\caption{Motivation of our work. (a) Illustration of class mismatch problem. (b) Strategy of existing approaches which utilizes the ID data, meanwhile down-weighting all the detected OOD data. (c) Our method utilizes the ID data as well as the recyclable OOD data, and then ignore the non-recyclable OOD data.}
	\label{fig:problem_illustration}\vspace{-3mm}
\end{figure}

Classical SSL includes graph-based methods~\cite{gong2014fick, gong2015saliency, gong2015deformed, gong2016multi, ke2020laplacian, li2020staged, mesgaran2020anisotropic, wan2021contrastive, wang2007label, zhou2004learning}, semi-supervised support vector machines~\cite{bennett1999semi}, co-training~\cite{blum1998combining, min2020mutually}. Recently, the research on SSL has made significant progress based on deep neural networks~\cite{lecun2015deep} with strong representation ability, and they usually utilize three types of training strategy to handle both labeled and unlabeled data, namely: 1) {\em entropy minimization}~\cite{grandvalet2005semi, lee2013pseudo} which impels networks to make confident predictions on unlabeled data; 2) {\em consistency regularization}~\cite{iscen2019label, laine2016temporal, li2020snowball, luo2018smooth, miyato2018virtual, tarvainen2017mean, zhang2019unsupervised, zhang2020wcp} which enforces that the perturbations on unlabeled data should not change their label predictions significantly; and 3) {\em data augmentation}~\cite{berthelot2019mixmatch, sohn2020fixmatch, wang2019semi, zhai2019s4l} which creates additional examples and label information to improve the generalizability of the learned classifier.

However, the above-mentioned SSL approaches rely on a basic assumption that the classes contained by labeled data (\textit{i.e.}, $\mathcal{C}_l$) and those contained by unlabeled data (\textit{i.e.}, $\mathcal{C}_u$) are the same, namely $ \mathcal{C}_l= \mathcal{C}_u$. Unfortunately, in real-world situations, such an assumption is difficult to satisfy as we actually do not know the labels of unlabeled data in advance. Such problem for realistic SSL is called {\em class mismatch}~\cite{oliver2018realistic} if some of the classes in unlabeled data are different from those in labeled examples, as shown in Figure~\ref{fig:problem_illustration}(a). Concretely, class mismatch means that the classes in labeled data $\mathcal{C}_l$ constitute a subset of the classes in unlabeled data $\mathcal{C}_u$, namely $\mathcal{C}_l\subseteq \mathcal{C}_u$ and $\mathcal{C}_u\setminus\mathcal{C}_l \ne \varnothing$. Our class mismatch definition follows~\cite{guo2020safe} which is quite practical: as unlabeled data are easy to acquire, so the classes of unlabeled data are likely to cover all the classes of labeled data. Here the unlabeled data that belong to the classes $\mathcal{C}_l$ are called {\em in-distribution} (ID) data, while the unlabeled data \textbf{only} belonging to $\mathcal{C}_u$ are called {\em out-of-distribution} (OOD) data. Due to the existence of OOD data, the traditional SSL methods will be confused and thus generating the degraded test performance regarding the interested classes in $\mathcal{C}_l$.

To solve the class mismatch problem, current works focus on leveraging ID data while trying to weaken the negative impact caused by the OOD data. For example, Chen \textit{et al.}~\cite{chen2020semi} propose a self-distillation method to filter out the probable OOD data based on their confidence scores. Guo \textit{et al.}~\cite{guo2020safe} utilize a weighting mechanism to decrease the influence of OOD data.  As shown in Figure~\ref{fig:problem_illustration} (b), both methods regard the rest detected OOD data as substantially harmful ones that should be discarded~\cite{chen2020semi} or down-weighted~\cite{guo2020safe}. However, here we argue that the detected OOD data are not completely useless. Some of them are informative and can actually be re-used in a suitable way to improve the classification performance. Considering an example, if ``dog'' and ``cat'' are two interested classes in $\mathcal{C}_l$ to be classified, and the examples of ``wolf'' and ``bird'' are private in $\mathcal{C}_u$ forming the OOD data. In this case, the unlabeled examples with the potential label ``wolf'' are much more useful than those of ``bird'' in classifying ``dog'' and ``cat'' in $\mathcal{C}_l$, as the appearances of wolf and dog are very similar. As a sequel, they can be activated for our classification task even though they are OOD data. In other words, the prior works~\cite{chen2020semi, guo2020safe} tackling class mismatch problem mostly deploy the selected ID data in an unlabeled set, while our method employs both ID data and some useful task-oriented OOD data to train an improved semi-supervised classifier. However, due to the distribution gap between ID data and OOD data, the detected OOD data cannot be directly used for network training, hence we perform a ``recycle'' procedure to properly re-use the OOD data, which is shown in Figure~\ref{fig:problem_illustration} (c). This is similar to garbage recycling in our daily life, where the useful part of the garbage can be recycled to create extra economic value. Here although the OOD data seem to be useless at the first glance, some of them are still beneficial for SSL training if they are properly used.

\begin{figure*}[t]
	\vspace{-4mm}
	\begin{center}
		\includegraphics[width=0.9\linewidth]{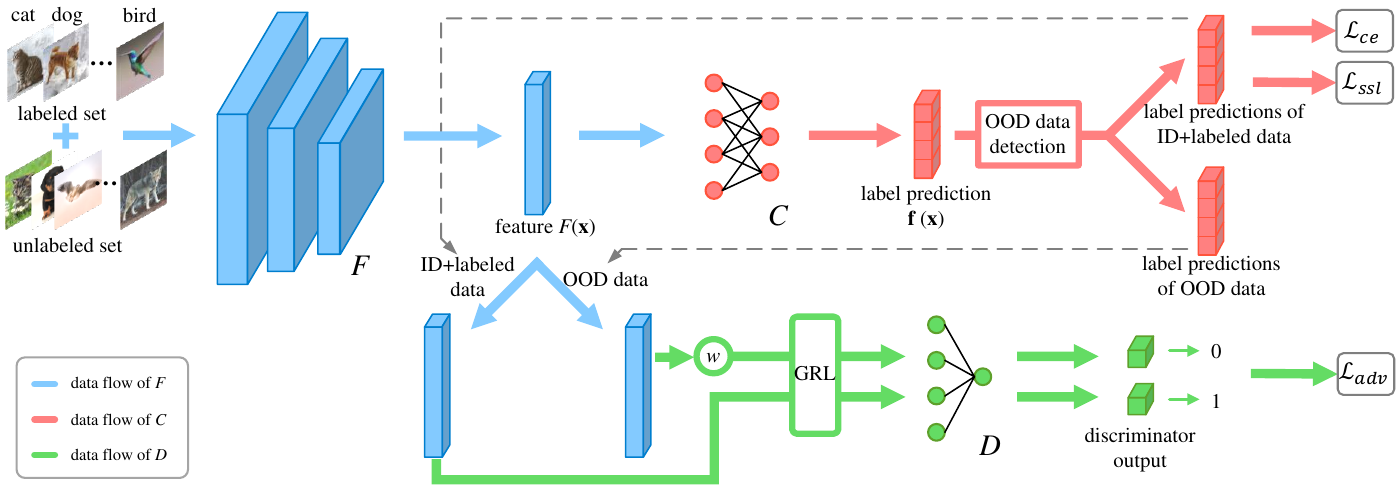}
	\end{center}
	\vspace{-4mm}
	\caption{The pipeline of our Transferable OOD data Recycling (TOOR) method, in which $F$, $C$ and $D$ denote the feature extractor, classifier, and discriminator, respectively. Given the labeled set and unlabeled set, $F$ extracts the feature $F(\mathbf{x})$ for each input image, which is then sent to $C$ to obtain the label prediction vector $\mathbf{f}(\mathbf{x})$ which encodes the probabilities of $\mathbf{x}$ to different classes. Based on $\mathbf{f}(\mathbf{x})$, the OOD data detection module decides whether $\mathbf{x}$ is an ID datum or an OOD datum. Then the determined ID data are used to compute $\mathcal{L}_{ce}$ and $\mathcal{L}_{ssl}$, and the determined OOD data are further decided to be recyclable or not. Based on the result of OOD data detection, we obtain the features of ID data and OOD data separately (see the gray dashed lines). Finally, the OOD data are weighted by the transferability score $w(\mathbf{x})$, and some of them are recycled by the adversarial learning process composed of the generator $F$ and discriminator $D$.  ``GRL'' denotes the gradient reversal layer~\cite{ganin2015unsupervised} that accomplishes adversarial learning. The above procedure including OOD data detection and adversarial learning for data recycling iterates until all ID data are found and transferable OOD data are recycled.}
	\vspace{-4mm}
	\label{fig:model}
\end{figure*}

Based on the above considerations, this paper proposes a ``Transferable OOD Data Recycling'' (TOOR) method for SSL under class mismatch which exploits the ID data and simultaneously recycles transferable OOD data in the unlabeled set for training a classifier. Specifically, TOOR divides the entire unlabeled set into three subsets in each training round, namely ID data, recyclable OOD data, and non-recyclable OOD data, where only the last subset is discarded for network training. Here ID data are automatically detected according to the fact that the softmax score of ID data would be higher than that of OOD data, while the recyclable OOD data and non-recyclable OOD data are decided by evaluating the transferability from them to the feature space of the labeled images. For recyclable OOD data, as their distribution still slightly differs from the labeled data, we propose to conduct domain adaptation through adversarial learning~\cite{cao2018partial, cao2019learning, ganin2016domain, tzeng2017adversarial, zhang2018importance} to reduce the potential gap between the two feature distributions. Hence, TOOR is able to learn on more examples than~\cite{chen2020semi, guo2020safe} since the transferable OOD data are also well adapted to the space of labeled data to aid SSL training. Thanks to the re-use of transferable OOD data, our TOOR method can be incorporated by many existing SSL approaches (\textit{e.g.}, Mean Teacher~\cite{tarvainen2017mean}, Virtual Adversarial Training (VAT)~\cite{miyato2018virtual}) so that class mismatch can be effectively tackled. Comprehensive experiments on typical real-world datasets (\textit{i.e.}, CIFAR10, SVHN, CIFAR100, and ImageNet) reveal that TOOR consistently outperforms other baseline methods in the presence of mismatched OOD data in the unlabeled set.

\vspace{-2mm}
\section{Related Work}
In this section, we first review some traditional SSL methods which do not consider the class mismatch, and then introduce recent representative class-mismatched SSL methodologies.

\vspace{-3mm}
\subsection{Traditional Semi-Supervised Learning}
Traditional SSL methods usually tackle the situation when the class sets of labeled data and unlabeled data are the same. For example, Pseudo-Labeling~\cite{lee2013pseudo} selects some confident predictions to generate hard labels for unlabeled data. Later, consistency-based methods conduct consistency training between temporally or spatially different models. For example, $\Pi$-Model~\cite{laine2016temporal, sajjadi2016regularization} utilizes two different models to make consistent predictions for perturbed image examples. Temporal Ensembling~\cite{laine2016temporal} generates a consistent learning target for each unlabeled data via using Exponential Moving Average (EMA), so that the historical network predictions can be memorized. Mean Teacher~\cite{tarvainen2017mean} also utilizes EMA to assemble a teacher model that contains historical knowledge of the student model and can better guide its learning process. After that, VAT~\cite{miyato2018virtual} computes adversarial perturbations which maximally change the unlabeled data, and then assigns pseudo labels to the perturbed data to enhance the model robustness.

Recent popular SSL methods mostly rely on data augmentation to improve the network generalizability. For instance, MixMatch~\cite{berthelot2019mixmatch} and ReMixMatch~\cite{berthelot2019remixmatch} employ the MixUp~\cite{zhang2017mixup} technique to augment the training data as well as the label information, which is beneficial to network training. FixMatch~\cite{sohn2020fixmatch} utilizes the label predictions of weakly augmented image data to guide the learning of strongly augmented image data and achieves state-of-the-art performance. However, as mentioned before, traditional SSL can hardly be applied to real-world problems as they cannot handle class mismatch problem which is widely observed in practice.

\vspace{-3mm}
\subsection{Class-Mismatched Semi-Supervised Learning}
Class-mismatched SSL methods consider the situation when the class sets of labeled data and unlabeled data are different. Such a problem is firstly raised by Laine \& Aila~\cite{laine2016temporal} and Oliver \textit{et al.}~\cite{oliver2018realistic}. After that, several works have been proposed to address this problem. For instance, Uncertainty Aware Self-Distillation (UASD)~\cite{chen2020semi} averages the historical predictions to detect the OOD data and proposes a self-distillation method to filter out the probable OOD data. Safe Deep Semi-Supervised Learning (DS$^3$L)~\cite{guo2020safe} deploys a meta-learning scheme to automatically down-weight the OOD data, which can decrease its negative impact. Multi-Task Curriculum Framework (MTCF)~\cite{yu2020multi} considers the ID data and OOD data as two different domains, and utilizes curriculum learning to distinguish them. The above methods have made some significant attempts towards solving the class mismatch problem. However, these methods all consider OOD data as harmful ones and thus failing to fully exploit their potential value.

\vspace{-4mm}
\section{The Proposed TOOR Approach}
\vspace{-1mm}
This section presents our proposed TOOR approach. In our class-mismatched SSL setting, we use the notations $\mathcal{X}$ and $\mathcal{Y}$ to denote the feature space and label space, respectively. Given a set of training image examples $\mathcal{D}=\left\{\mathbf{x}_i\in\mathcal{X}\subset\mathbb{R}^d,i=1,2,\cdots,n,n=l+u\ \mbox{with}\ l\ll u\right\}$ in which the first $l$ image examples are labeled with $\left\{y_i\right\}_{i=1}^l\in\mathcal{Y}=\left\{1,2,\cdots,c\right\}$ where $c$ is the number of known classes, and the remaining $u$ image examples are unlabeled. We use $\mathcal{D}_l=\left\{(\mathbf{x}_1, y_1), (\mathbf{x}_2, y_2),\cdots,(\mathbf{x}_l,y_l)\right\}$ to denote the labeled set drawn from the joint distribution $P_{\mathcal{X}\times \mathcal{Y}}$ defined on $\mathcal{X}\times\mathcal{Y}$, and $\mathcal{D}_u=\left\{\mathbf{x}_{l+1}, \mathbf{x}_{l+2}, \cdots, \mathbf{x}_{l+u}\right\}$ to represent the unlabeled set that is sampled from marginal distribution $P_{\mathcal{X}}$. Note that the marginal distribution $P_{\mathcal{X}}$ of labeled data and unlabeled data are the same. However, $\mathcal{D}_u$ is assumed to be composed of an ID dataset $\mathcal{D}_{id}$ whose labels are in $\mathcal{Y}$ and an OOD dataset $\mathcal{D}_{ood}$ whose labels are in $\mathcal{Y}^{\prime}$ with typically $\mathcal{Y} \subset \mathcal{Y}^{\prime}$, \textit{i.e.}, $\mathcal{D}_u=\mathcal{D}_{id}\cup\mathcal{D}_{ood}$. As a result, there is a distribution gap exists between the true joint distributions of labeled data $P_{\mathcal{X}\times\mathcal{Y}}$ and unlabeled data $P_{\mathcal{X}\times\mathcal{Y}^{\prime}}$.

The main goal of TOOR is to effectively utilize the class-mismatched training set $\mathcal{D}=\mathcal{D}_l \cup \mathcal{D}_u$ to find a semi-supervised classifier that can properly leverage $\mathcal{D}_u$ so that any unseen image $\mathbf{x}$ with unknown label $y\in\mathcal{Y}$ can be correctly classified. The model of TOOR can be concisely formulated as:\vspace{-3mm}
\begin{equation}
	\begin{aligned}
		\min_{\theta_F, \theta_C}\max_{\theta_D} \ & \underbrace{\frac{1}{l}\sum_{i=1}^{l}\mathcal{L}_{ce}(\mathbf{x}_i, y_i; \theta_F, \theta_C)}_{\text{supervised fidelity term}}\\+&\underbrace{\lambda \frac{1}{l+\left|\mathcal{D}_{id}\right|}\sum_{\mathbf{x}_i\in\mathcal{D}_{id}\cup\mathcal{D}_l}\mathcal{L}_{ssl}(\mathbf{x}_i;\theta_F, \theta_C)}_{\text{ID data exploration term}}\\+& \underbrace{\gamma\frac{1}{\left|\mathcal{D}_{ood}\right|}\sum_{\mathbf{x}_i\in\mathcal{D}_{ood}}w(\mathbf{x}_i)\cdot\mathcal{L}_{adv}(\mathbf{x}_i; \theta_F, \theta_D)}_{\text{OOD data recycling term}},\label{general}
	\end{aligned}
\end{equation}
in which $F$ is an image feature extractor, $C$ is a classifier, $D$ is a discriminator, and $\theta_F$, $\theta_C$, and $\theta_D$ are their parameters, respectively. The notation ``$\left|\cdot\right|$'' denotes the size of the corresponding set. In Eq.~\eqref{general}, the first term is dubbed {\em supervised fidelity term} which involves the standard cross-entropy loss $\mathcal{L}_{ce}(\cdot)$ to compare the network prediction $C(F(\mathbf{x}_i))$ on every labeled image and its ground-truth label $y_i$. The second term refers to {\em ID data exploration term} in which $\mathcal{L}_{ssl}(\cdot)$ denotes the loss defined on ID data and it can be any regularizer in existing SSL method such as consistency regularizer~\cite{laine2016temporal, tarvainen2017mean} or manifold regularizer~\cite{belkin2006manifold, geng2012ensemble, yu2019tangent}. The third term is dubbed {\em OOD data recycling term} which introduces an adversarial learning loss $\mathcal{L}_{adv}(\cdot)$ to ``recycle'' the transferable OOD data. Here the OOD data are found by examining their transferability score $w(\mathbf{x}_i)$ which will be detailed in Section ~\ref{recycling}. Through such a recycling procedure, our TOOR approach can maximally exploit class-mismatched datasets without including useless or harmful OOD data, and meanwhile re-use the rich information contained by the transferable unlabeled image examples, leading to superior performance to other methods. The parameters $\lambda$ and $\gamma$ are nonnegative coefficients that control the relative weights of the above three terms.

The general procedure of the proposed TOOR algorithm is shown in Figure~\ref{fig:model}. Given the labeled image set $\mathcal{D}_l$ and unlabeled image set $\mathcal{D}_u$, we use the feature extractor $F$ to compute the feature representations $F(\mathbf{x})$ for $\mathbf{x}\in \mathcal{D}_l\cup \mathcal{D}_u$. Then, a classifier $C$ is imposed on $F(\mathbf{x})$ to obtain the label prediction vector $\mathbf{f}(\mathbf{x})$ for each of the input images. Based on $\mathbf{f}(\mathbf{x})$, the ID data can be found which are used to compute $\mathcal{L}_{ssl}$ together with the labeled data. The decided OOD data are further sent to the adversarial learning branch so that the recyclable OOD data are called back and the non-recyclable OOD data are completely discarded. Specifically, all OOD data weighted with the transferability scores are combined with the ID data to act as a generator, and they are employed to confuse the discriminator $D$. Then $D$ should try its best to distinguish the presented data as ID data (\textit{i.e.}, 1) or OOD data (\textit{i.e.}, 0). During the iterative process between OOD data detection and adversarial learning for recycling, our detected ID dataset expands from the initial limited labeled image set by gradually absorbing the considered ID images and the transferable OOD images. From the above explanations on the procedure of our TOOR, we see that OOD data detection, adversarial learning for recycling, and semi-supervised training parts are critical in our method, and they will be detailed in Sections ~\ref{OOD detection}, ~\ref{recycling}, and ~\ref{SSL}, respectively.

\vspace{-3mm}
\subsection{OOD Data Detection}\label{OOD detection}
{\em OOD data detection} aims to correctly distinguish the unlabeled image data into  ID data and OOD data. Many existing works~\cite{hinton2015distilling, liang2017enhancing} have shown that it can be accomplished by investigating the {softmax scores} of unlabeled data during network training. Specifically, given an input image $\mathbf{x}$, its label prediction $\mathbf{f}(\mathbf{x})$ output by the classifier is a $c$-dimensional vector $\left[f_1(\mathbf{x}),f_2(\mathbf{x}),\cdots,f_c(\mathbf{x})\right]^\top$, where $\left\{f_i(\mathbf{x})\right\}_{i=1}^c$ can be interpreted as the probability that $\mathbf{x}$ belongs to class $i$. We follow~\cite{liang2017enhancing} to achieve the scaled label prediction $\mathbf{S}(\mathbf{x}; \tau) = \left[S_1(\mathbf{x}; \tau), S_2(\mathbf{x}; \tau), \cdots, S_c(\mathbf{x}; \tau)\right]^\top$, where $\left\{S_i(\mathbf{x}; \tau)\right\}_{i=1}^c$ are computed as\vspace{-2mm}
\begin{equation}
	S_i(\mathbf{x}; \tau) = \frac{\exp{(f_i(\mathbf{x})/\tau)}}{\sum_{j=1}^{c}\exp{(f_j(\mathbf{x})/\tau)}},\label{softmax}
\end{equation}
in which $\tau \in \mathbb{R}^+$ is a temperature scaling parameter~\cite {hinton2015distilling, liang2017enhancing} that controls the concentration level of the distribution. Here the maximum value of the elements in $\mathbf{S}(\mathbf{x}; \tau)$ is dubbed {\em softmax score}~\cite{liang2017enhancing}, which is computed as
\begin{equation}
	s(\mathbf{x})= \max_{i\in\left\{1,\cdots,c\right\}}{S_i(\mathbf{x}; \tau)}.\label{maxscore}
\end{equation}
It has been demonstrated in~\cite{liang2017enhancing} that the softmax scores of ID data are significantly larger than those of OOD data, and thus the softmax scores of different examples can be utilized to judge the unlabeled images as ID data or OOD data.

However, in our TOOR approach, due to the aforementioned gradual expansion of ID dataset during the training process, the softmax scores of certain OOD data may oscillate, which make their results of OOD data detection not consistent across successive iterations. To address this problem, we conduct temporal ensembling~\cite{laine2016temporal} before the computation of softmax scores to achieve stabilized predictions on all unlabeled data. Such stabilization can achieve better results than the averagely assembled scores as done in~\cite{chen2020semi} which may introduce noise from the earlier stage of network training. Concretely, we assemble the label predictions of unlabeled data from historical iterations via using EMA, which assigns greater weights to recent predictions while exponentially decreasing the weights of early predictions. Therefore, the assembled label prediction of $\mathbf{x}$ is computed as
\begin{equation}
	\hat{\mathbf{S}}(\mathbf{x}; \tau)^{(t)}=\eta \hat{\mathbf{S}}(\mathbf{x}; \tau)^{(t-1)}+(1-\eta)\mathbf{S}(\mathbf{x}; \tau)^{(t)},\label{tempens}
\end{equation}
where $\mathbf{S}(\mathbf{x}; \tau)^{(t)}$ denotes the scaled label prediction whose elements are computed according to Eq.~\eqref{softmax} at the $t$-th iteration; $\hat{\mathbf{S}}(\mathbf{x}; \tau)^{(t)}$ and $\hat{\mathbf{S}}(\mathbf{x}; \tau)^{(t-1)}$ denote the assembled label predictions at the $t$-th iteration and the $(t-1)$-th iteration, respectively. The coefficient $\eta \in \left[0, 1\right]$ is a momentum parameter that decides how far the ensemble reaches into training history. Such assembled label prediction $\hat{\mathbf{S}}(\mathbf{x}; \tau)$ varies smoothly and will not be significantly changed across different iterations, so it can be used to compute a stabilized softmax score $\hat{s}(\mathbf{x})$ through a similar computation as Eq.~\eqref{maxscore}. Formally, we denote $\hat{s}(\mathbf{x}) = \max_{i\in\left\{1,\cdots,c\right\}}{\hat{S}_i(\mathbf{x}; \tau)}$, where $\hat{S}_i(\mathbf{x}; \tau)$ represents the $i$-th element of $\hat{\mathbf{S}}(\mathbf{x}; \tau)$. As a result, the stabilized score enables our method to make consistent identification on OOD data.

Given the stabilized softmax score $\hat{s}(\mathbf{x})$, we use an OOD threshold $\delta$ to separate OOD data from ID data in the unlabeled set $\mathcal{D}_u$. Specifically, an image is considered as an ID datum if its stabilized softmax score is larger than $\delta$, and an OOD datum otherwise. Detailed explanation on how $\delta$ is chosen is deferred to Section~\ref{performance}. By introducing  $t(\mathbf{x}; \delta)$ as an indication variable regarding $\mathbf{x}$, this process is formulated as
\begin{equation}
	t(\mathbf{x};\delta)=\left\{ 
	\begin{aligned}
		&1, \ \text{if}\ \hat{s}(\mathbf{x})>\delta \\
		&0, \ \text{if}\ \hat{s}(\mathbf{x})\le\delta
	\end{aligned}
	\right..\label{OOD detection eq}
\end{equation}
That is to say, the images $\mathbf{x}$ with $t(\mathbf{x};\delta)=1$ are determined as ID data and they will be incorporated by $\mathcal{D}_{id}$ to compute the ID data exploration term in Eq.~\eqref{general}. The experimental results in Section~\ref{performance} show that the computed score can lead to impressive performance. 

\vspace{-3mm}
\subsection{Adversarial Learning for Recycling}\label{recycling}
After detecting the OOD data as mentioned in the above subsection, common approach would discard the OOD data, however, we argue that their potential value should not be completely ignored. Specifically, we hope to find the recyclable images in $\mathcal{D}_{ood}$ and then transfer them to the space of $\mathcal{D}_l\cup\mathcal{D}_{id}$ such that their contained information can fully be extracted for training a semi-supervised classifier. To this end, we treat $\mathcal{D}_{ood}$ as source distribution and $\mathcal{D}_l\cup\mathcal{D}_{id}$ as target distribution, and propose to leverage adversarial domain adaptation technique to mitigate the distribution gap. Here we treat $\mathcal{D}_l\cup\mathcal{D}_{id}$ rather than $\mathcal{D}_l$ as target distribution as $\mathcal{D}_l$ contains very limited labeled data which cannot faithfully reflect the corresponding distribution and may cause poor generalizability of the obtained classifier. 

Adversarial domain adaptation~\cite{cao2018partial, cao2019learning, ganin2016domain, tzeng2017adversarial, zhang2018importance} aims to learn class discriminative and domain invariant features by using adversarial learning~\cite{goodfellow2014generative}. In our problem, adversarial learning can help mitigate the joint distribution shift between labeled data and unlabeled data. Moreover, it can be utilized to help explore the value of transferable OOD data. Specifically, the parameters $\theta_D$ of the discriminator $D$ are learned to distinguish the previously identified OOD data from ID data by minimizing a cross-entropy loss. Meanwhile, the parameters $\theta_F$ of the feature extractor $F$ are learned to deceive the discriminator by maximizing the same cross-entropy loss. In this way, the domain shift between ID data and transferable OOD data is closed, in which the OOD data that are ``easily'' transferred are likely to be recycled. Note that although the original classes of OOD data are different from ID data, some recyclable OOD data still contain discriminative information as they are close to one specific ID classes and dissimilar to all the rest ID class. Therefore, conducting adversarial training on the recyclable OOD data encourages the model to produce confident predictions on some uncertain ID data, thus improving the model robustness. Hence, by using $\mathcal{D}_l\cup\mathcal{D}_{id}$ as the transfer target, the above adversarial process can be formulated as the min-max game below:
\begin{equation}
	\begin{aligned}
		\min_{\theta_F}\max_{\theta_D}\mathcal{L}_{adv}&=\frac{1}{\left|\mathcal{D}_{ood}\right|}\sum_{\mathbf{x}_i\in \mathcal{D}_{ood}}w(\mathbf{x}_i)\log D(F(\mathbf{x}_i))\\&+\frac{1}{\left|\mathcal{D}_{id}\right|+l}\sum_{\mathbf{x}_i\in\mathcal{D}_{id}\cup\mathcal{D}_l}\log(1-D(F(\mathbf{x}_i))),\label{minmax}
	\end{aligned}
\end{equation}
where $w(\mathbf{x}_i)$ is the transferability score that helps to find recyclable OOD data, and the computation of this score will be detailed later. Through such a min-max optimization procedure, the two adversarial opponents converge to a situation where the features of transferable OOD data will be pushed near to the labeled data and ID data, so as to fool the discriminator. Note that we also train the classifier $C$ by minimizing the supervised cross-entropy loss $\mathcal{L}_{ce}$ on the original labeled data, which is simultaneously conducted with adversarial learning. Hence, we can successfully extract helpful knowledge from OOD data for our classification task on the interested label space $\mathcal{Y}$.

Here we propose to utilize two cues to find the potential transferable OOD data from the unlabeled set $\mathcal{D}_u$.  Firstly, if the discriminator $D$ cannot tell whether an image $\mathbf{x}_i\in\mathcal{D}_u$ is from the source or target domain, we know that $\mathbf{x}_i$ is quite ambiguous as its representation is close to both ID data and OOD data. Therefore, it is likely to be a transferable image example that should be recycled. To depict this, we introduce a {\em domain similarity score} $w_d(\mathbf{x}_i)$ for $\mathbf{x}_i$, which has been widely used in many domain adaptation methods~\cite{cao2018partial, cao2019learning, you2020Universal}. Secondly, if the classifier $C$ attributes $\mathbf{x}_i$ to a certain class $y\in\mathcal{Y}$ with a strong tendency, we learn that $\mathbf{x}_i$ probably belongs to this interested class, so it should be recycled. To describe this, we introduce a {\em class tendency score} $w_c(\mathbf{x}_i)$, which can effectively extract the discriminative information during semi-supervised training. By adaptively integrating $w_d(\mathbf{x}_i)$ and $w_c(\mathbf{x}_i)$, we acquire the transferability score $w(\mathbf{x}_i)$ for any $\mathbf{x}_i\in\mathcal{D}_u$, which serves as the weight for $\mathbf{x}_i$ to perform the min-max game in Eq.~\eqref{minmax}. In the following, we explain the computations for $w_d(\mathbf{x}_i)$, $w_c(\mathbf{x}_i)$ and the integrated $w(\mathbf{x}_i)$.

\textbf{Domain similarity score $w_d(\mathbf{x}_i)$.} Our discriminator $D$ is trained to distinguish the ID data from OOD data. The output of discriminator can be interpreted as
\begin{equation}
	D(F(\mathbf{x}_i))=p(\mathbf{x}_i\in\mathcal{D}_{id}\mid\mathbf{x}_i), \  \mathbf{x}_i\in\mathcal{D}_u,\label{prob}
\end{equation}
where $p(\cdot)$ denotes probability in this paper. Eq.~\eqref{prob} means that the output value of domain discriminator provides the likelihood of an OOD example $\mathbf{x}_i$ belonging to the domain of $\mathcal{D}_{id}$. Consequently, if $D(F(\mathbf{x}_i))$ is large, we know that $\mathbf{x}_i$ is similar to the known space of $\mathcal{D}_l\cup\mathcal{D}_{id}$. As a result, we should properly recycle these examples by assigning them large scores. On the other hand, if $D(F(\mathbf{x}_i))$ is small, the corresponding $\mathbf{x}_i$ might not come from the space of $\mathcal{D}_l\cup\mathcal{D}_{id}$. These images should have small scores such that both the discriminator and the classifier will ignore them. Hence, the score reflecting domain information is formulated as
\begin{equation}
	\tilde{w}_{d}(\mathbf{x}_i) = D(F(\mathbf{x}_i)).
\end{equation}
To obtain more discriminative score assignments across OOD data, we normalize the scores of all unlabeled data to compute domain similarity score, which is
\begin{equation}
	w_{d}(\mathbf{x}_i) = \frac{\tilde{w}_{d}(\mathbf{x}_i)}{\frac{1}{u}\sum_{j=1}^{u}\tilde{w}_{d}(\mathbf{x}_j)},\ \mathbf{x}_i\in\mathcal{D}_{u}.\label{norm}
\end{equation}
The above normalization equipped with the global normalizer (\textit{i.e.}, the denominator in Eq~\eqref{norm}) helps to enlarge the scores for the potential recyclable OOD data, and meanwhile decreasing the scores of non-transferable OOD data to prevent them from being recycled.

\textbf{Class tendency score $w_c(\mathbf{x}_i)$.} Apart from the domain similarity score, the assembled label predictions of OOD data generated by $C$ also contain rich transferability information. Concretely, the assembled label predictions of ID data can provide valuable clue in evaluating the transferability since the classifier $C$ has been trained on labeled set $\mathcal{D}_{l}$ and thus possessing considerable discriminability. Hence, inspired by~\cite{yao2020deep}, we employ the predictive margin between the largest and the second-largest elements of assembled label prediction vector $\hat{\mathbf{S}}(\mathbf{x}; \tau)$ to establish class tendency score, which is computed as
\begin{equation}
	\tilde{w}_{c}(\mathbf{x}_i)=\max_{j\in\left\{1,\cdots,c\right\}}{\hat{S}_j(\mathbf{x}; \tau)} - \max_{k\in\left\{1,\cdots,c\right\};\ k\ne j}{\hat{S}_k(\mathbf{x}; \tau)}.
\end{equation}
If the predictive margin of an OOD example is large, it implies that the example has relatively large tendency to one certain category, then we consider such OOD examples as transferable data and they could be recycled to the corresponding class $j$. On the other hand, if the margin is small, which means that the label of this example is quite unclear, it should be excluded for recycling. Note that another method such as~\cite{you2020Universal} utilizes entropy to incorporate label information, but it is not suitable in SSL as the entropy of unlabeled data is usually minimized through SSL regularization, thus making the SSL model being overconfident on some OOD data. However, our class tendency score only focuses on the margin of the two largest prediction probabilities to fully extract the class discriminative information, thus avoiding the overconfident problem inherited by entropy minimization. Similar to the operation on domain similarity score $w_d(\mathbf{x}_i)$, here we also normalize $w_c(\mathbf{x}_i)$ as Eq.~\eqref{norm}, which is formulated as
\begin{equation}
	w_{c}(\mathbf{x}_i) = \frac{\tilde{w}_{c}(\mathbf{x}_i)}{\frac{1}{u}\sum_{j=1}^u\tilde{w}_{c}(\mathbf{x}_j)},\ \mathbf{x}_i\in\mathcal{D}_{u}.\label{norm2}
\end{equation}

\textbf{Adaptive integration of $w_d(\mathbf{x}_i)$ and $w_c(\mathbf{x}_i)$.} To calculate the transferability scores of all OOD data, we need to combine the obtained domain similarity scores and class tendency scores in a proper way. Specifically, these two scores should be explicitly weighted to yield good performance. By employing the vectors $\mathbf{w}_d=[w_d(\mathbf{x}_1),\cdots,w_d(\mathbf{x}_{\left|\mathcal{D}_{ood}\right|})]^{\top}$ and $\mathbf{w}_c=[w_c(\mathbf{x}_1),\cdots,w_c(\mathbf{x}_{\left|\mathcal{D}_{ood}\right|})]^{\top}$ to encode the domain similarity scores and class tendency scores for all $\mathbf{x}_i\in \mathcal{D}_{ood}$, here we propose to utilize their variances to compute the tradeoff weights. Concretely, if the variance of $w_d(\mathbf{x}_i)$ or $w_c(\mathbf{x}_i)$ is large, which means that the values of contained elements are discriminative for characterizing the transferability of all $\mathbf{x}_i\in\mathcal{D}_{ood}$, then it should be paid more attention in composing the final transferability score $w(\mathbf{x}_i)$. In contrast, if the variance of $w_d(\mathbf{x}_i)$ or $w_c(\mathbf{x}_i)$ is small, then it is less helpful in evaluating the transferability of $\mathbf{x}_i$, so its contribution in computing $w(\mathbf{x}_i)$ should be suppressed. Mathematically, we have the following convex combination:
\begin{equation}
	\vspace{-2mm}
	\begin{aligned}
		w(\mathbf{x}_i)&=\frac{\mathrm{var}(\mathbf{w}_d)}{\mathrm{var}(\mathbf{w}_d)+\mathrm{var}(\mathbf{w}_c)}w_d(\mathbf{x}_i)\\&+\frac{\mathrm{var}(\mathbf{w}_c)}{\mathrm{var}(\mathbf{w}_d)+\mathrm{var}(\mathbf{w}_c)}w_c(\mathbf{x}_i),
	\end{aligned}\label{adaptive weighting}
\end{equation}
where $\mathrm{var}(\cdot)$ denotes the variance computation regarding the input vector. Since we have re-scaled the range of these two scores $w_d(\mathbf{x}_i)$ and $w_c(\mathbf{x}_i)$ to the same level as in Eqs.~\eqref{norm} and \eqref{norm2}, the variances of them can be directly employed to compare their importance in composing the transferability score $w(\mathbf{x}_i)$. Finally, we can find the transferable OOD data by weighing each OOD datum with $w(\mathbf{x}_i)$ and then perform the weighted min-max game as Eq.~\eqref{minmax}.

After the transferable OOD data have been recycled by the above weighted min-max game, their feature representations will fall into the feature space of $\mathcal{D}_l\cup\mathcal{D}_{id}$, so that they will act as ID data. Furthermore, they will be included in SSL training to improve the performance of our classification task on the interested classes.

\renewcommand{\algorithmicrequire}{ \textbf{Input:}}
\renewcommand{\algorithmicensure}{ \textbf{Output:}}
\setlength{\textfloatsep}{10pt}
\begin{algorithm}[t]
	\scriptsize
	\caption{Training process for our TOOR method}
	\label{alg1}
	\begin{algorithmic}[1]
		\REQUIRE{Labeled set $\mathcal{D}_l=\left\{(\mathbf{x}_1, \mathrm{y}_1),\cdots,(\mathbf{x}_l,\mathrm{y}_l)\right\}$, class-mismatched unlabeled set $\mathcal{D}_u=\left\{\mathbf{x}_{l+1}, \cdots, \mathbf{x}_{l+u}\right\}$.}
		\STATE Train feature extractor $F$ and classifier $C$ on labeled set $\mathcal{D}_l$ by minimizing the supervised fidelity term in Eq.~\eqref{general};\\
		\FOR {$i=1$ to MaxIter}
		\STATE Compute assembled label prediction $\hat{\mathbf{S}}(\mathbf{x}; \tau)$ according to Eq.~\eqref{tempens} and the stabilized softmax score $\hat{s}(\mathbf{x}) = \max_{i\in\left\{1,\cdots,c\right\}}{\hat{S}_i(\mathbf{x}; \tau)}$;\\
		\STATE Perform OOD data detection to find the ID dataset $\mathcal{D}_{id}$ and OOD dataset $\mathcal{D}_{ood}$ through Eq.~\eqref{OOD detection eq};\\
		\STATE Weigh each OOD datum $\mathbf{x}$ with the transferability score $w(\mathbf{x})$ computed through Eq.~\eqref{adaptive weighting};\\
		\STATE OOD data recycling and network training by minimizing Eq.~\eqref{general}.\\
		\ENDFOR
		\ENSURE Discriminator parameter $\theta_D$; and SSL model with parameters $\theta_F$ and $\theta_C$ for classification.
	\end{algorithmic}
\end{algorithm}

\vspace{-3mm}
\subsection{Semi-Supervised Training}\label{SSL}
Through the aforementioned OOD data detection and adversarial learning for recycling, we can take full advantage of class-mismatched datasets by finding unlabeled ID data and recyclable OOD data, as well as filtering out the useless non-recyclable OOD data. Then, we can utilize the useful original labeled data, ID data, and recyclable OOD data to implement semi-supervised training by deploying any existing SSL regularizer to the $\mathcal{L}_{ssl}$ in Eq.~\eqref{general}, such as consistency loss in ~\cite{laine2016temporal, lee2013pseudo, tarvainen2017mean}, virtual adversarial training loss in~\cite{miyato2018virtual}, entropy minimization in~\cite{grandvalet2005semi}, and so on.

Therefore, the general framework of the TOOR algorithm can be instantiated by substituting a specified SSL regularizer $\mathcal{L}_{ssl}$ into the ID data exploration term in Eq.~\eqref{general}, and meanwhile replacing the OOD data recycling term in Eq.~\eqref{general} with the weighted min-max game in Eq.~\eqref{minmax}. Our TOOR method is summarized in Algorithm~\ref{alg1}. Later experiments in Section~\ref{singleset} will show that TOOR can enhance the performances of many typical SSL methods in handling the class mismatch problem.

\vspace{-2mm}
\section{Experiments}\label{experiments}
In this section, we conduct exhaustive experiments to validate the proposed TOOR approach. Firstly, we provide implementation details of our TOOR method (Section~\ref{implementation}). Then we evaluate the performance of our method under class mismatch on single datasets (Section~\ref{singleset}). Furthermore, we evaluate the capability of TOOR under a more challenging case when the labeled and unlabeled data come from different datasets with overlapped classes (Section~\ref{crossset}). Finally, a detailed performance study will be provided to verify the effectiveness of OOD data detection and recycling procedure in our TOOR approach (Section~\ref{performance}).

\begin{table}[H]
	\vspace{-4mm}
	\scriptsize
	\centering
	\setlength{\tabcolsep}{1.2mm}
	\caption{Architecture of backbone network $F$.}
	\vspace{-2mm}
	\label{backbone}
	\begin{threeparttable}
		\begin{tabular}{ccc}
			\toprule[1.5pt]
			\textbf{Group Name}& \textbf{Layer}  & \textbf{Hyperparameters} \\
			\midrule[1pt]
			&	Input&	32$\times$32 RGB image\\
			&	Translation&	Randomly $\{\Delta x, \Delta y\} \sim \left[-2, 2\right]$\\
			&	Horizontal flip\tnote{*}&	Randomly $p = 0.5$\\
			&	Gaussian noise&	$\sigma=0.15$\\
			\midrule[0.6pt]
			conv1&	Convolutional&	16 filters, conv3$\times$3\\
			\midrule[0.6pt]
			\multirow{6}*{conv2}&
			
			\multirow{6}*{
				$\begin{bmatrix}
					\text{Batch-Normalization}\\
					\text{Leaky-ReLU}\\
					\text{Convolutional}\\
					\text{Batch-Normalization}\\
					\text{Leaky-ReLU}\\
					\text{Convolutional}\\
				\end{bmatrix}\times4$
			}
			&$momentum=1\times10^{-3}$\\
			&&$negative\_slope=0.1$\\
			&&32 filters, conv3$\times$3\\
			&&$momentum=1\times10^{-3}$\\
			&&$negative\_slope=0.1$\\
			&&32 filters, conv3$\times$3\\
			\midrule[0.6pt]
			\multirow{6}*{conv3}&
			
			\multirow{6}*{
				$\begin{bmatrix}
					\text{Batch-Normalization}\\
					\text{Leaky-ReLU}\\
					\text{Convolutional}\\
					\text{Batch-Normalization}\\
					\text{Leaky-ReLU}\\
					\text{Convolutional}\\
				\end{bmatrix}\times4$
			}
			&$momentum=1\times10^{-3}$\\
			&&$negative\_slope=0.1$\\
			&&64 filters, conv3$\times$3\\
			&&$momentum=1\times10^{-3}$\\
			&&$negative\_slope=0.1$\\
			&&64 filters, conv3$\times$3\\
			\midrule[0.6pt]
			\multirow{6}*{conv4}&
			
			\multirow{6}*{
				$\begin{bmatrix}
					\text{Batch-Normalization}\\
					\text{Leaky-ReLU}\\
					\text{Convolutional}\\
					\text{Batch-Normalization}\\
					\text{Leaky-ReLU}\\
					\text{Convolutional}\\
				\end{bmatrix}\times4$
			}
			&$momentum=1\times10^{-3}$\\
			&&$negative\_slope=0.1$\\
			&&128 filters, conv3$\times$3\\
			&&$momentum=1\times10^{-3}$\\
			&&$negative\_slope=0.1$\\
			&&128 filters, conv3$\times$3\\
			\midrule[0.6pt]
			\multirow{3}*{avg-pool}&
			Batch-Normalization&	$momentum=1\times10^{-3}$\\
			&Leaky-ReLU&		$negative\_slope=0.1$\\
			&Average Pooling&	$output size=1$\\
			\midrule[0.6pt]
			&	Fully Connected&	$classes=6$\\
			&	Softmax \\
			\bottomrule[1.5pt]
		\end{tabular}
		\begin{tablenotes}
			\footnotesize
			\item[*] \scriptsize Not applied on SVHN experiments.
		\end{tablenotes}  
	\end{threeparttable}
\vspace{-2mm}
\end{table}

\begin{table}[H]
	\vspace{-6mm}
	\scriptsize
	\centering
	\caption{Architecture of our discriminator $D$.}
	\vspace{-2mm}
	\label{discriminator}
	\begin{threeparttable}
		\begin{tabular}{ccc}
			\toprule[1.5pt]
			\textbf{Layer}&	\textbf{Hyperparameters} \\
			\midrule[1pt]
			GRL&	flip-coefficient\\
			Linear&		128$\rightarrow$1,024\\
			ReLU&	\\
			Dropout&	$p=0.5$\\
			Linear&		1,024$\rightarrow$1,024\\
			ReLU&	\\
			Dropout&	$p=0.5$\\
			Linear&		1,024$\rightarrow$1\\
			Sigmoid&	\\
			\bottomrule[1.5pt]
		\end{tabular}
	\end{threeparttable}
\vspace{-2mm}
\end{table}

\begin{figure*}
	\vspace{-6mm}
	\begin{center}
		\includegraphics[width=\linewidth]{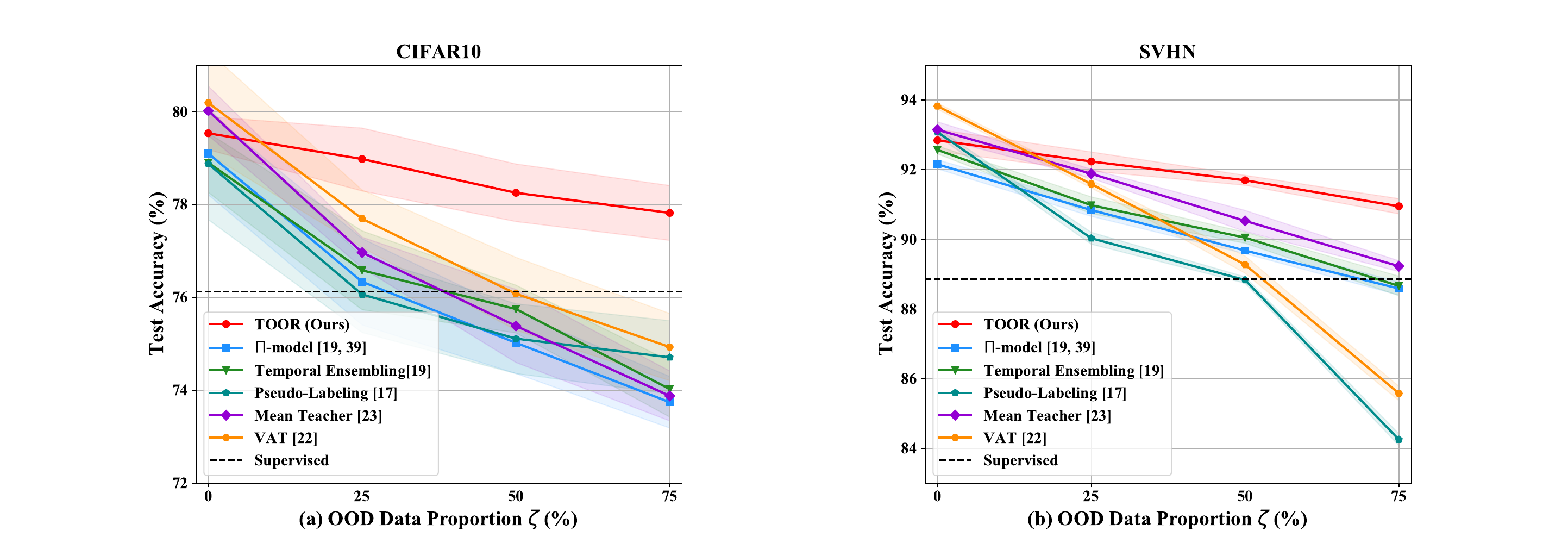}
	\end{center}
\vspace{-6mm}
	\caption{Test accuracies of traditional SSL methods under class mismatch with varied $\zeta$. (a) On CIFAR10 dataset. (b) On SVHN dataset. The shaded regions with the curves indicate the standard deviations of the accuracies over five runs.}
	\label{fig:sslcompare}
	\vspace{-3mm}
\end{figure*}

\vspace{-6mm}
\subsection{Implementation details}\label{implementation}
Our TOOR is implemented by using the batch size of 100 and is trained for 500,000 iterations. The network training is conducted by Adam optimizer~\cite{kingma2014adam} with the weight decay factor 0.2 after 400,000 iterations. All our experiments can be conducted on a single P40 GPU, and it takes 8 hours to train the proposed TOOR method. For other compared baseline methods, it took 12 hours to train DS$^3$L~\cite{guo2020safe}, 6 hours to train UASD~\cite{chen2020semi}, and 7 hours to train MTCF~\cite{yu2020multi}. Next we present the configurations of our TOOR by specifying the network architectures as well as the hyperparameter values.

\textbf{Backbone network $F$.} To be consistent with previous works, we choose Wide ResNet-28-2~\cite{zagoruyko2016wide} as our backbone network $F$, which is exactly the same as the ones used in~\cite{oliver2018realistic, guo2020safe}, and we consider the last fully connected layer followed by a softmax operation as the classifier $C$. The architecture of the adopted Wide ResNet-28-2 is shown in Table~\ref{backbone}. Note that the ``Horizontal flip'' is not applied in SVHN as such dataset is simple that ``Horizontal flip'' does not bring further performance improvement, therefore it is not needed. Moreover, to achieve a fair comparison, we implement all baseline methods with the same backbone architecture.

\textbf{Discriminator $D$.}	We also choose the same structure as in~\cite{cao2019learning} to construct our discriminator, which is shown in Table~\ref{discriminator}. The flip-coefficient $flip\_coeff$ in the GRL~\cite{ganin2015unsupervised} aims to suppress the noisy signals from the discriminator at the early stages of training procedure, which ramps up from 0 to 1 by following the function $flip\_coeff=\frac{2}{1+\exp(-10\times (iter-pretrain\_iter) \times \frac{1}{400,000})}-1$, where $iter$ denotes the current training iteration, $pretrain\_iter$ denotes the number of iterations for supervised training and is set to 5,000 in our experiments. Note that some domain adaptation methods~\cite{cao2019learning, zhang2018importance, you2020Universal} employ an extra non-adversarial discriminator to produce transferability. However, in class mismatch problem, we find using a non-adversarial discriminator does not bring further performance improvement, hence we only use one discriminator to avoid complicating our method.

\textbf{Hyperparameters.} We set the temperature $\tau$ in Eq.~\eqref{softmax} as 0.8 to compute the softmax scores, and set the OOD data detection threshold $\delta$ as 0.95 for SVHN and 0.9 for other datasets to distinguish the ID data from OOD data. The trade-off parameters $\lambda$ and $\gamma$ ramp up from 0 to 1 by following the functions $\lambda=\exp(-5\times{(1-\min(\frac{iter}{200,000}, 1))}^2)$ and $\gamma=\exp(-5\times{(1-\min(\frac{iter}{400,000}, 1))}^2)$, respectively. Since our method is a general framework which can be applied to many traditional SSL methods, we use different initial learning rates for our backbone network by following the original settings of the adopted specific SSL methods. For discriminator $D$, we set the initial learning rate to 0.001.

\begin{figure*}
	\vspace{-6mm}
	\begin{center}
		\includegraphics[width=\linewidth]{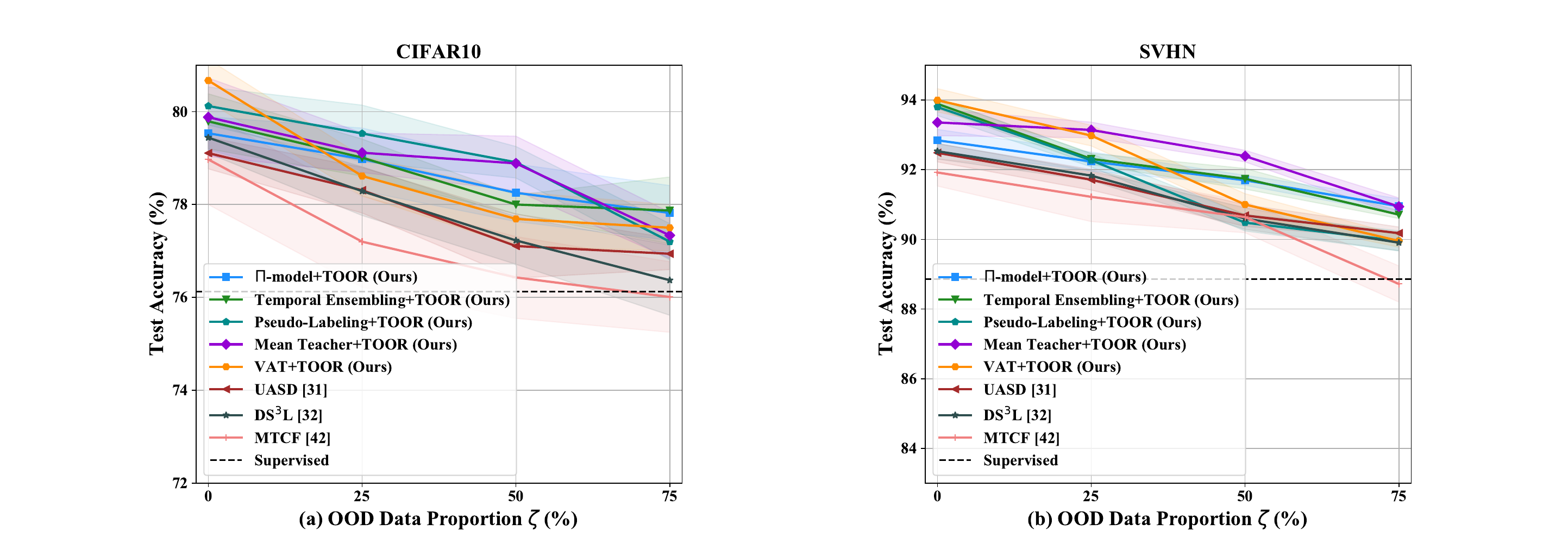}
	\end{center}
	\vspace{-6mm}
	\caption{Test accuracies of various class-mismatched SSL methods under class mismatch with varied $\zeta$. (a) On CIFAR10 dataset. (b) On SVHN dataset. The shaded regions with the curves indicate the standard deviations of the accuracies over five runs.}
	\label{fig:cmcompare}
	\vspace{-3mm}
\end{figure*}

\vspace{-4mm}
\subsection{Evaluation on Single-Dataset Scenario}\label{singleset}
In this subsection, we create a class mismatch case between labeled and unlabeled data in each single dataset. Specifically, we use CIFAR10~\cite{krizhevsky2009learning} and SVHN~\cite{netzer2011reading} datasets to construct two class-mismatched datasets, respectively. Specifically, \textbf{CIFAR10} contains 50,000 and 10,000 natural images with the size of 32$\times$32 for training and test accordingly, which consist of six animal classes (\textit{e.g.}, ``bird'', ``cat'', ``deer'', ``dog'', ``frog'', and ``horse'') and four transportation tool classes (\textit{e.g.}, ``airplane'', ``automobile'', ``ship'', and ``truck''). Here we follow~\cite{guo2020safe} by randomly choosing 400 images from each of the six animal classes in the training set to construct the labeled set, and picking up 20,000 training images from all ten classes to compose the unlabeled set. In this way, the images belonging to the animal classes are ID data and those from the transportation tool classes are OOD data. \textbf{SVHN} is composed of 73,257 training images and 26,032 test images with the resolution of 32$\times$32, which are collected from real-world house numbers. This dataset contains ten classes, namely the ten digits ``0''$\sim$``9''. We randomly choose 100 images from the each of six classes ``0''$\sim$``5'' in the training set to compose the labeled set, and randomly sample 20,000 training images from all ten classes ``0''$\sim$``9'' to form the unlabeled set.

\subsubsection{Comparison with Traditional SSL Methods}\label{comparessl}
To testify the capability of TOOR for class-mismatched SSL tasks, we vary the proportion of OOD data in unlabeled set (denoted as $\zeta$ hereinafter) to investigate the performance of our method under different numbers of class-mismatched data. The five traditional SSL methods for comparison include Pseudo-Labeling~\cite{lee2013pseudo}, $\Pi$-model~\cite{laine2016temporal, sajjadi2016regularization}, Temporal Ensembling~\cite{laine2016temporal}, VAT~\cite{miyato2018virtual}, and Mean Teacher~\cite{tarvainen2017mean}. Furthermore, we also train the backbone network Wide ResNet-28-2 on the labeled set to form the ``Supervised'' baseline. We are aware that there is another trend of SSL methods relying on data augmentation such as MixMatch~\cite{berthelot2019mixmatch} and FixMatch~\cite{sohn2020fixmatch}, which is orthogonal to the main contribution of this work, so here they are not included for comparison. Concretely, we set $\zeta$ to $\{0\%, 25\%, 50\%, 75\%\}$ to observe the performances of TOOR and baseline methods, where $\zeta=0\%$ means that there are no OOD data. For TOOR, we adopt the consistency regularization used in $\Pi$-model as the SSL regularizer $\mathcal{L}_{ssl}$.

For conducting traditional SSL methods, we choose the hyperparameters by following~\cite{oliver2018realistic}, which have been carefully tuned to achieve their best performances. Specifically, for Pseudo-Labeling~\cite{lee2013pseudo}, we set the pseudo label threshold as 0.95. For Temporal Ensembling~\cite{laine2016temporal}, the EMA factor is set to 0.6. For VAT~\cite{miyato2018virtual}, the perturbation magnitude is tuned to 1.0 for SVHN dataset and 6.0 for CIFAR10 dataset, such different choice of magnitude is because that the objects contained by CIFAR10 images is more complex than SVHN images, hence a stronger perturbation is imposed to the images. Regarding Mean Teacher~\cite{tarvainen2017mean}, the EMA factor is adjusted to 0.95.

For both CIFAR10 and SVHN, we randomly select OOD data for five times under each $\zeta$ to establish the training set, and report the average test accuracies as well as the standard deviations of comparators over five independent runs. The experimental results on both datasets are shown in Figure~\ref{fig:sslcompare}. We can see that the performances of all baseline methods seriously degrade when $\zeta$ increases, as the gradually incorporated OOD data significantly mislead the training of the above SSL methods. In contrast, our method shows more stable performance than other SSL methods in both datasets, and constantly surpasses the supervised baseline, which demonstrates that TOOR successfully avoids the possible confusion brought by the OOD data in mismatched classes. Specifically, when $\zeta=75\%$, our TOOR can still achieve 77.81$\%$ and 90.95$\%$ test accuracies on CIFAR10 and SVHN, respectively, which are significantly higher than 74.93$\%$ and 89.23$\%$ achieved by the second best method.

\subsubsection{Comparison with Class-Mismatched SSL Methods}\label{comparecm}
In this part, we apply the proposed TOOR framework to some typical SSL methods (\textit{e.g.}, Pseudo-Labeling~\cite{lee2013pseudo}, $\Pi$-model~\cite{laine2016temporal, sajjadi2016regularization}, Temporal Ensembling~\cite{laine2016temporal}, VAT~\cite{miyato2018virtual}, and Mean Teacher~\cite{tarvainen2017mean}) to enable them to handle class-mismatched cases, and compare them with three existing approaches (\textit{i.e.}, UASD~\cite{chen2020semi}, DS$^3$L~\cite{guo2020safe}, and MTCF~\cite{yu2020multi}) that deal with class-mismatched SSL problem.


The parameters of baseline methods have been carefully tuned to achieve their best performances. Particularly, for UASD, we choose the ensemble size as 10 to integrate historical predictions. For DS$^3$L, we use the same weighting network structure as the original paper. The initial learning rate for the weighting network of DS$^3$L is set to 0.001.

Similar to the above experiment, we also investigate the results of various methods under $\zeta=\{0\%, 25\%, 50\%, 75\%\}$, and the average accuracies of five runs on CIFAR10 and SVHN datasets are shown in Figure~\ref{fig:cmcompare}. We can see that all the five traditional SSL methods combined with TOOR (\textit{i.e.}, ``$\Pi$-model+TOOR'', ``Temporal Ensembling+TOOR'', ``Pseudo-Labeling+TOOR'', ``Mean Teacher+TOOR'', and ``VAT+TOOR'') perform satisfactorily, and the significant performance drop revealed by Figure~\ref{fig:sslcompare} does not appear anymore, which indicates that TOOR is quite general and can help many traditional SSL methods to tackle the class mismatch problem. Moreover, the five SSL methods enhanced by TOOR outperform UASD, DS$^3$L, and MTCF in most cases, which again demonstrates the effectiveness of our method. Specifically, when $\zeta=50\%$ which means that half of the unlabeled images are OOD data, our ``Mean teacher + TOOR'' achieves the accuracy of 78.88$\%$, which is much higher than UASD, DS$^3$L, and MTCF with the accuracies 77.10$\%$, 77.22$\%$, and 76.43$\%$ correspondingly on CIFAR10 dataset. On SVHN dataset, the accuracy of ``Mean teacher + TOOR'' is as high as 92.39$\%$, which outperforms UASD, DS$^3$L, and MTCF with the accuracies 90.68$\%$, 90.59$\%$, and 90.64$\%$ accordingly.

\begin{table}[t]
	\vspace{-2mm}
	\small
	\centering
	\setlength{\tabcolsep}{1mm}
	\caption{Test accuracies (\%) with standard deviations over five runs on the constructed CIFAR100+ImageNet dataset. The best results are highlighted in \textbf{bold}.}
	\label{crossset_table}
	\vspace{-3mm}
	\begin{threeparttable}
		\begin{tabular}{lc}
			\toprule[1pt]
			Method&  CIFAR100+ImageNet \\
			\midrule[0.6pt]
			Supervised& $45.31\pm1.12$  \\
			\midrule[0.6pt]
			Pseudo-Labeling~\cite{lee2013pseudo}&  $43.90\pm0.46$  \\
			$\Pi$-model~\cite{laine2016temporal, sajjadi2016regularization}& $43.81\pm1.28$  \\
			Temporal Ensembling~\cite{laine2016temporal}&  $44.10\pm0.31$  \\
			VAT~\cite{miyato2018virtual}& $44.49\pm0.69$  \\
			Mean Teacher~\cite{tarvainen2017mean}&  $43.23\pm0.84$  \\
			UASD~\cite{chen2020semi}& $44.90\pm0.47$  \\
			DS$^3$L~\cite{guo2020safe}& $45.10\pm1.25$  \\
			MTCF~\cite{yu2020multi}&  $46.34\pm0.81$  \\
			\midrule[0.6pt]
			TOOR (Ours)&  $\bm{49.15}\pm\textbf{0.76}$  \\
			\bottomrule[1pt]
		\end{tabular}
	\end{threeparttable}
\end{table}
\vspace{-3mm}

\begin{table*}[]
	\vspace{-3mm}
	\scriptsize
	\centering
	\setlength{\tabcolsep}{1.5mm}
	\caption{\scriptsize Test accuracies (\%) with standard deviations over five runs when varying the number of labeled examples. The numbers of unlabeled examples in all datasets are set to 20000. The best results are highlighted in \textbf{bold}.}
	\label{varyinglabel_table}
	\vspace{-3mm}
	\begin{tabular}{cccccc}
		\toprule[1pt]
		&      & \multicolumn{4}{c}{\# labeled data}                                                                           \\ 
		\midrule[0.6pt]
		&      & 300 (50 for each class)   & 600 (100 for each class)  & 900 (150 for each class)  & 1200 (300 for each class) \\ 
		\midrule[0.6pt]
		\multirow{4}{*}{SVHN}              & UASD~\cite{chen2020semi} & $86.28\pm0.36$&	$90.68\pm0.31$&	$91.72\pm0.29$&	$92.96\pm0.41$\\
		& DS$^3$L~\cite{guo2020safe} &$86.42\pm0.41$	&$90.59\pm0.31$	&$91.20\pm0.64$	&$93.12\pm0.48$\\
		& MTCF~\cite{yu2020multi} &$85.91\pm0.27$	&$90.64\pm0.45$	&$90.87\pm0.39$	&$92.16\pm0.55$\\
		& TOOR &$\bm{87.73}\pm\bm{0.20}$	&$\bm{91.69}\pm\bm{0.14}$	&$\bm{92.68}\pm\bm{0.18}$	&$\bm{94.05}\pm\bm{0.12}$\\ 
		\midrule[0.6pt]
		&      & 1800 (300 for each class) & 2400 (400 for each class) & 3000 (500 for each class) & 3600 (600 for each class) \\ 
		\midrule[0.6pt]
		\multirow{4}{*}{CIFAR10}           & UASD~\cite{chen2020semi} &$74.50\pm0.64$	&$77.10\pm0.69$	&$78.10\pm0.82$	&$79.66\pm0.96$\\
		& DS$^3$L~\cite{guo2020safe} &$74.56\pm0.73$	&$77.22\pm0.52$	&$78.29\pm0.62$	&$80.01\pm0.60$\\
		& MTCF~\cite{yu2020multi} &$74.11\pm0.39$	&$76.43\pm0.88$	&$77.98\pm0.59$	&$79.12\pm0.82$\\
		& TOOR &$\bm{75.77}\pm\bm{0.59}$	&$\bm{78.25}\pm\bm{0.62}$	&$\bm{79.66}\pm\bm{0.81}$	&$\bm{81.38}\pm\bm{0.54}$ \\ 
		\midrule[0.6pt]
		&      & 4800 (80 for each class)  & 6000 (100 for each class) & 7200 (120 for each class) & 8400 (140 for each class) \\ 
		\midrule[0.6pt]
		\multirow{4}{*}{CIFAR100+ImageNet} & UASD~\cite{chen2020semi} &$42.07\pm0.58$	&$44.90\pm0.47$ &$46.38\pm0.79$	&$48.20\pm0.40$\\
		& DS$^3$L~\cite{guo2020safe} &$43.99\pm0.54$	&$45.10\pm1.25$	&$47.11\pm0.73$	&$48.96\pm0.63$\\
		& MTCF~\cite{yu2020multi} &$45.69\pm0.74$	&$46.34\pm0.81$	&$48.92\pm0.58$	&$50.05\pm0.77$\\
		& TOOR &$\bm{47.19}\pm\bm{0.83}$	&$\bm{49.15}\pm\bm{0.76}$	&$\bm{51.34}\pm\bm{0.65}$	&$\bm{52.98}\pm\bm{0.79}$\\ 
		\bottomrule[1pt]
	\end{tabular}
\vspace{-2mm}
\end{table*}

\begin{table*}[]
	\scriptsize
	\centering
	\setlength{\tabcolsep}{1.0mm}
	\caption{\scriptsize Test accuracies (\%) with standard deviations over five runs when varying the number of unlabeled examples. The numbers of labeled examples in SVHN, CIFAR10, and CIFAR100+ImageNet are set to 600, 2400, and 6000, respectively. The best results are highlighted in \textbf{bold}.}
	\label{varyingunlabel_table}
	\vspace{-3mm}
	\begin{tabular}{cccccc}
		\toprule[1pt]
		&      & \multicolumn{4}{c}{\# unlabeled data}                                                                           \\ 
		\midrule[0.6pt]
		&      & 15000 (1500 for each class)&	20000 (2000 for each class)&	25000 (2500 for each class)&	30000 (3000 for each class) \\ 
		\midrule[0.6pt]
		\multirow{4}{*}{SVHN}              & UASD~\cite{chen2020semi} &$89.02\pm0.28$	&$90.68\pm0.31$	&$91.33\pm0.07$	&$92.62\pm0.24$\\
		& DS$^3$L~\cite{guo2020safe} &$89.14\pm0.30$	&$90.59\pm0.31$	&$91.51\pm0.29$	&$92.98\pm0.19$\\
		& MTCF~\cite{yu2020multi} &$\bm{90.91}\pm\bm{0.21}$	&$90.64\pm0.45$	&$91.34\pm0.57$	&$93.02\pm0.44$\\
		& TOOR &$90.85\pm0.17$	&$\bm{91.69}\pm\bm{0.14}$	&$\bm{92.25}\pm\bm{0.18}$	&$\bm{93.39}\pm\bm{0.12}$\\ 
		\midrule[0.6pt]
		&      & 15000 (1500 for each class)&	20000 (2000 for each class)&	25000 (2500 for each class)&	30000 (3000 for each class) \\ 
		\midrule[0.6pt]
		\multirow{4}{*}{CIFAR10}           & UASD~\cite{chen2020semi} &$76.23\pm0.46$	&$77.10\pm0.69$	&$77.97\pm0.50$	&$78.20\pm0.52$\\
		& DS$^3$L~\cite{guo2020safe} &$76.00\pm0.49$	&$77.22\pm0.52$	&$78.10\pm0.45$	&$78.23\pm0.66$\\
		& MTCF~\cite{yu2020multi} &$76.12\pm0.72$	&$76.43\pm0.88$	&$77.59\pm0.63$	&$78.09\pm0.77$\\
		& TOOR &$\bm{77.83}\pm\bm{0.75}$	&$\bm{78.25}\pm\bm{0.62}$	&$\bm{78.98}\pm\bm{0.82}$	&$\bm{80.46}\pm\bm{0.88}$\\ 
		\midrule[0.6pt]
		&      & 15000 (150 for each class)&	20000 (200 for each class)&	25000 (250 for each class)&	30000 (300 for each class) \\ 
		\midrule[0.6pt]
		\multirow{4}{*}{CIFAR100+ImageNet} & UASD~\cite{chen2020semi} &$44.21\pm0.38$	&$44.90\pm0.47$ &$45.92\pm0.68$	&$46.36\pm0.73$\\
		& DS$^3$L~\cite{guo2020safe} &$44.87\pm0.43$	&$45.10\pm1.25$	&$46.14\pm0.71$	&$47.33\pm0.90$\\
		& MTCF~\cite{yu2020multi} &$45.46\pm0.59$	&$46.34\pm0.81$	&$47.68\pm1.17$	&$48.99\pm1.55$\\
		& TOOR &$\bm{48.19}\pm\bm{0.92}$	&$\bm{49.15}\pm\bm{0.76}$	&$\bm{49.99}\pm\bm{1.13}$	&$\bm{50.10}\pm\bm{1.08}$\\ 
		\bottomrule[1pt]
	\end{tabular}
\vspace{-4mm}
\end{table*}

\subsection{Evaluation on Cross-Dataset Scenario}\label{crossset}
To further test the effectiveness of our proposed method, we create a more challenging scenario when labeled and unlabeled data come from two different datasets with overlapped classes, which contains a much larger distribution gap between labeled and unlabeled data than the single-dataset scenario conducted in Section~\ref{singleset}. In our experiments, we choose CIFAR100~\cite{krizhevsky2009learning} and ImageNet~\cite{chrabaszcz2017downsampled} to form our labeled and unlabeled set, respectively. \textbf{CIFAR100} has the same image data as CIFAR10 but contains 100 classes. \textbf{ImageNet} contains 1,331,167 images from 1,000 classes. To create the dataset for evaluation, we choose 6,000 images from 60 classes in CIFAR100 as the labeled set where the selected 60 classes can also be found in ImageNet. Then we sample 20,000 images from 100 classes in ImageNet to form the unlabeled set, which contains the 60 classes that correspond to the chosen classes in the labeled set. Besides, the remaining 40 classes in the unlabeled set are randomly selected from the rest 940 classes in ImageNet. Here we denote the established dataset as ``\textbf{CIFAR100+ImageNet}'', in which the detailed class mapping from CIFAR100 to ImageNet is presented in the \textbf{appendix}. Note that in~\cite{chen2020semi, yu2020multi}, TinyImageNet is used to create class mismatch from CIFAR100, but it only contains 200 classes, which do not contain all the classes that exist in CIFAR100, so TinyImageNet does not satisfy our problem setting. Fortunately, ImageNet can extend TinyImageNet to 1,000 classes and contains all the classes in CIFAR100, which is ideal for our experiments.

For all experiments conducted in this subsection, we set the OOD proportion $\zeta=50\%$ and adopt $\Pi$-model as the backbone method. The experimental results are shown in Table~\ref{crossset_table}. We can see that TOOR significantly outperforms all the other compared methods on the constructed dataset, which indicates the capability of TOOR in tackling challenging classification task under large class mismatch. Specifically, TOOR significantly surpasses the supervised baseline with 3.84$\%$ on the averaged test accuracy while almost all the other compared methods show performance degradation than the supervised baseline. This is due to the large distribution shift between labeled and unlabeled data, which makes the original unlabeled ID data harmful to the performance of SSL. Thanks to the introduced transferring strategy, the proposed TOOR method can successfully eliminate the distribution gap via using adversarial learning, therefore achieving an improved performance than the supervised baseline.

\vspace{-2mm}
\subsection{Experiment on Different Numbers of Examples}\label{varying_number}
To testify the robustness of our proposed method under the varied number of labeled and unlabeled examples, we adopt consistency regularization used in $\Pi$-model as the SSL regularizer $\mathcal{L}_{ssl}$ and set the OOD proportion $\zeta=50\%$ to compare our method with three class-mismatched SSL methods including UASD~\cite{chen2020semi}, DS$^3$L~\cite{guo2020safe}, and MTCF~\cite{yu2020multi} on SVHN, CIFAR10, and CIFAR100+ImageNet datasets.

\textbf{Varying Number of Labeled Examples.} We fix the number of unlabeled data points to 20,000 and only change the number of labeled examples. The experimental results are shown in Table~\ref{varyinglabel_table}. We can see that our method can outperform all the compared baseline methods in all considered situations, which indicates the robustness of our method when presented with different numbers of labeled examples.

\textbf{Varying Number of Unlabeled Examples.} We fix the number of labeled examples in each dataset, and only change the number of unlabeled examples. We show the experimental results in Table~\ref{varyingunlabel_table}. We can see that our method can still outperform the compared baseline methods in most cases, indicating the effectiveness of TOOR in dealing with different numbers of unlabeled data points.

\begin{figure}[t]
	\begin{center}
		\includegraphics[width=\linewidth]{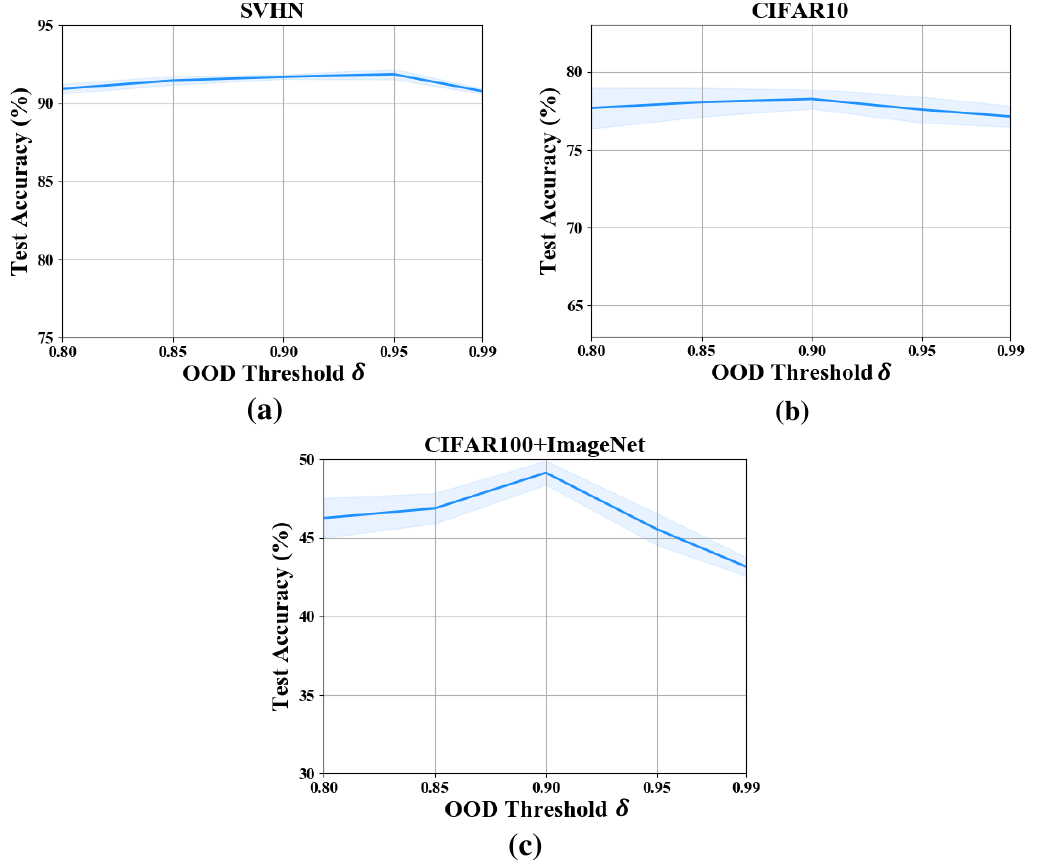}
	\end{center}
\vspace{-4mm}
	\caption{Parameter sensitivity analysis on OOD threshold $\delta$. The curves with shaded regions indicate the accuracies with standard deviation over three runs. (a) results on SVHN dataset. (b) results on CIFAR10 dataset. (c) results on CIFAR100+ImageNet dataset.}
	\label{fig:paramsens}
\end{figure}

\vspace{-2mm}
\subsection{Performance Study}\label{performance}
From the above experimental results presented in Sections~\ref{singleset} and \ref{crossset}, we can see that TOOR achieves very encouraging results. Here we further analyze the effects of the key components in TOOR, and study the behind reasons for TOOR in achieving good performance. Specifically, we see that the OOD data detection in Section~\ref{OOD detection} and transferable OOD data recycling in Section~\ref{recycling} are critical to tackling the class mismatch problem and boosting the performance, so next we validate their effectiveness on various datasets. For all the experiments in this section, we set the OOD proportion $\zeta=50\%$ and adopt $\Pi$-model as the backbone method.

\textbf{OOD data detection.} As mentioned in Section~\ref{OOD detection}, the performance of OOD data detection largely depends on the OOD threshold and the stabilized softmax scores. Firstly, to testify the effect of different values of OOD threshold $\delta$, we conduct parameter sensitivity analysis on the OOD threshold $\delta$ on different datasets including SVHN, CIFAR10, and CIFAR100+ImageNet. The experimental results are presented in Figure~\ref{fig:paramsens}. We can find that when $\delta$ increases from 0.8 to 0.9, the learning accuracies increase on all three datasets, this is due to the harmful OOD data are correctly filtered out. However, when $\delta$ increases from 0.9 to 0.99, the learning performances would drop. This is because that most of the ID data, as well as the transferable OOD data, are erroneously left out from training. As a result, the generalization ability is greatly limited, and thus causing performance degradation. Therefore, we set the threshold $\delta$ to 0.95 for SVHN dataset and to 0.9 for other datasets. We keep such threshold setting in all later experiments.

\begin{figure}[t]
	\begin{center}
		\includegraphics[width=\linewidth]{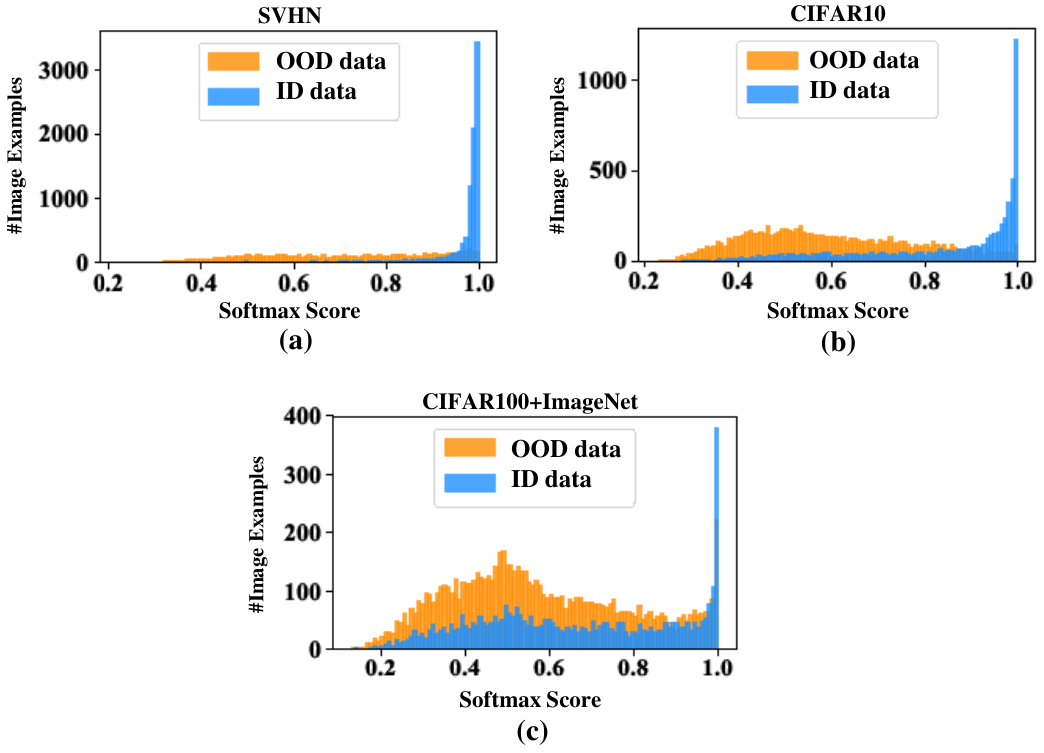}
	\end{center}
	\vspace{-4mm}
	\caption{The histograms of the computed softmax scores $\hat{s}(\mathbf{x})$ over all unlabeled images, where (a) denotes SVHN dataset, (b) denotes CIFAR10 dataset, and (c) denotes CIFAR100+ImageNet dataset.}
	\label{fig:performance}
\end{figure}

Moreover, to show the reasonability of our stabilized softmax scores, we plot the scores of ID data and OOD data from SVHN, CIFAR10, and CIFAR100+ImageNet datasets in Figures~\ref{fig:performance} (a), (b), and (c), respectively. In the SVHN and CIFAR10 datasets, it is noteworthy that most of the ID data have scores close to 1, and the scores of OOD data approximately show a uniform distribution as most of them are left out from the network training. Therefore, the computed softmax score provides valuable information in discriminating ID data and OOD data. As for the CIFAR100+ImageNet dataset, we can see that most of the scores of ID data are still larger than the scores of OOD data in such a challenging scenario, hence the computed softmax score can still be a satisfactory criterion.

\begin{table}[t]
	\vspace{-2mm}
	\scriptsize
	\centering
	\setlength{\tabcolsep}{0.5mm}
	\caption{Mean values with standard deviations of the non-stabilized softmax scores of the detected ID data, recyclable OOD data, and the non-recyclable OOD data.}
	\label{nonstabilized_softmax_score_table}
	\begin{threeparttable}
		\begin{tabular}{cccc}
			\toprule[1pt]
			& ID data  & Recyclable OOD data & Non-recyclable OOD data \\
			\midrule[0.6pt]
			SVHN& $0.98\pm0.03$ & $0.94\pm0.08$ & $0.72\pm0.16$ \\
			CIFAR10& $0.95\pm0.07$ & $0.87\pm0.13$ & $0.64\pm0.25$ \\
			CIFAR100+ImageNet& $0.78\pm0.20$ & $0.72\pm0.21$ & $0.58\pm0.29$ \\
			\bottomrule[1pt]
		\end{tabular}
	\end{threeparttable}
\end{table}

Furthermore, to give a quantitative evidence that our OOD data detection can identify the recyclable OOD data and meanwhile avoid over-detecting the non-recyclable OOD data, we set the OOD threshold $\delta$ to 0.95 for SVHN and 0.9 for other datasets, and provide the non-stabilized softmax scores of the detected ID data, recyclable OOD data, and the non-recyclable OOD data. The experiments are conducted in three independent runs and the mean values with standard deviations of the results are shown in Table~\ref{nonstabilized_softmax_score_table}. We can see that the non-stabilized softmax scores of recyclable OOD data are close to the scores of ID data, and the softmax scores of non-recyclable OOD data are kept relatively low compared with recyclable OOD data, which indicates that our OOD data detection can correctly identify ID data as well as recyclable OOD data, and meanwhile avoid over-detecting too much harmful non-recyclable OOD data.

\begin{figure}[t]
	\vspace{-3mm}
	\begin{center}
		\includegraphics[width=1.05\linewidth]{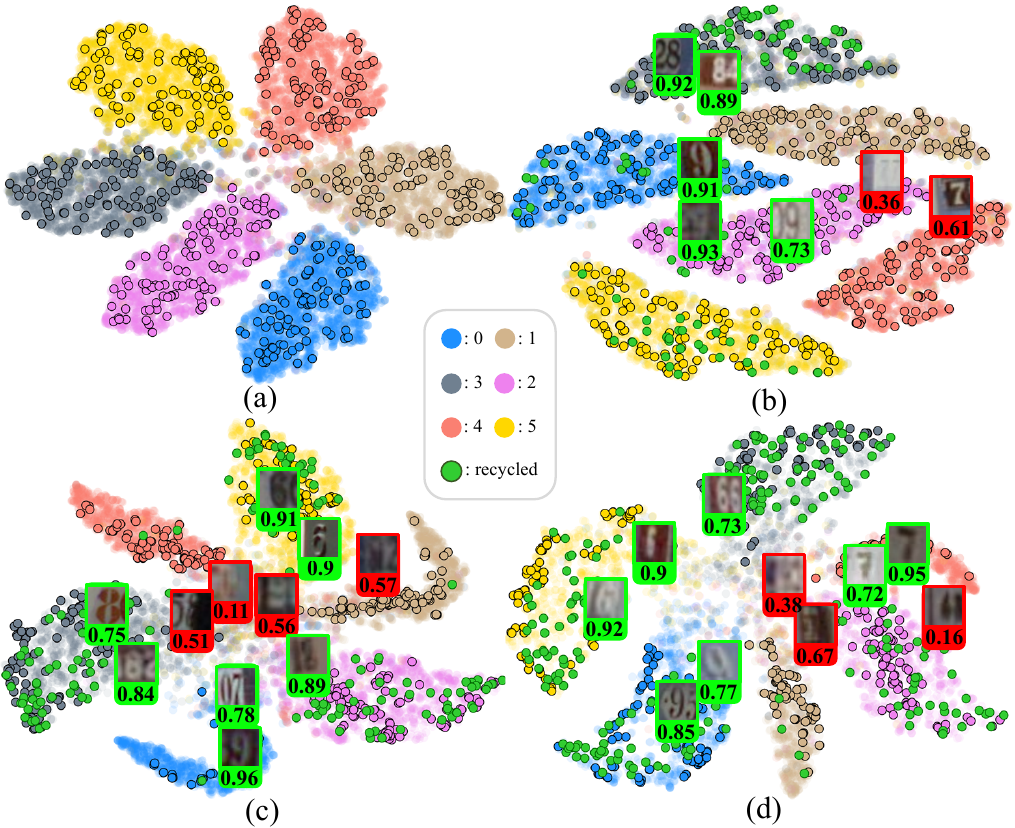}
	\end{center}
	\caption{t-SNE visualizations of the image representations. Figures (a)$\sim$(d) correspond to the results obtained on SVHN with $\zeta$ equaling 0\%, 25\%, 50\%, 75\%, respectively. Different colors denote the image data from different classes, where the green dots indicate the recycled OOD data determined by TOOR, and the dots with black boundary denote the originally labeled data. Besides, some of the recyclable OOD images (green boxes) and non-recyclable OOD images (red boxes) are visualized, of which the transferability scores $w(\mathbf{x})$ are also indicated below the images.}
	\label{fig:tsne}
\end{figure}

\begin{table}[t]
	\centering
	\setlength{\tabcolsep}{0.5mm}
	\caption{Percentages (\%) of the recycled OOD data. Averaged values with standard deviations are provided.}
	\vspace{-2mm}
	\label{recycle_percentage_table}
	\begin{threeparttable}
		\begin{tabular}{cc}
			\toprule[1pt]
			& Recycled OOD data  \\
			\midrule[0.6pt]
			SVHN& $34.45\pm0.65$ \\
			CIFAR10&  $29.73\pm1.10$ \\
			CIFAR100+ImageNet& $38.90\pm1.85$ \\
			\bottomrule[1pt]
		\end{tabular}
	\end{threeparttable}
\end{table}

\textbf{OOD data recycling.}
To further study the contribution of the recycling procedure, we conduct experiments on various datasets to show how many OOD data are recycled and incorporated into networking training. The results including the mean percentages with standard deviations over three independent runs are shown in Table~\ref{recycle_percentage_table}. We can see that there are a relatively small proportion of OOD data are recycled and incorporated into network training, and most of the OOD data are still regarded as harmful parts that should be left out, which means our recycling process is quite selective and would not be influenced by too much harmful OOD data.


\begin{figure}[t]
	\vspace{-3mm}
	\begin{center}
		\includegraphics[width=1.05\linewidth]{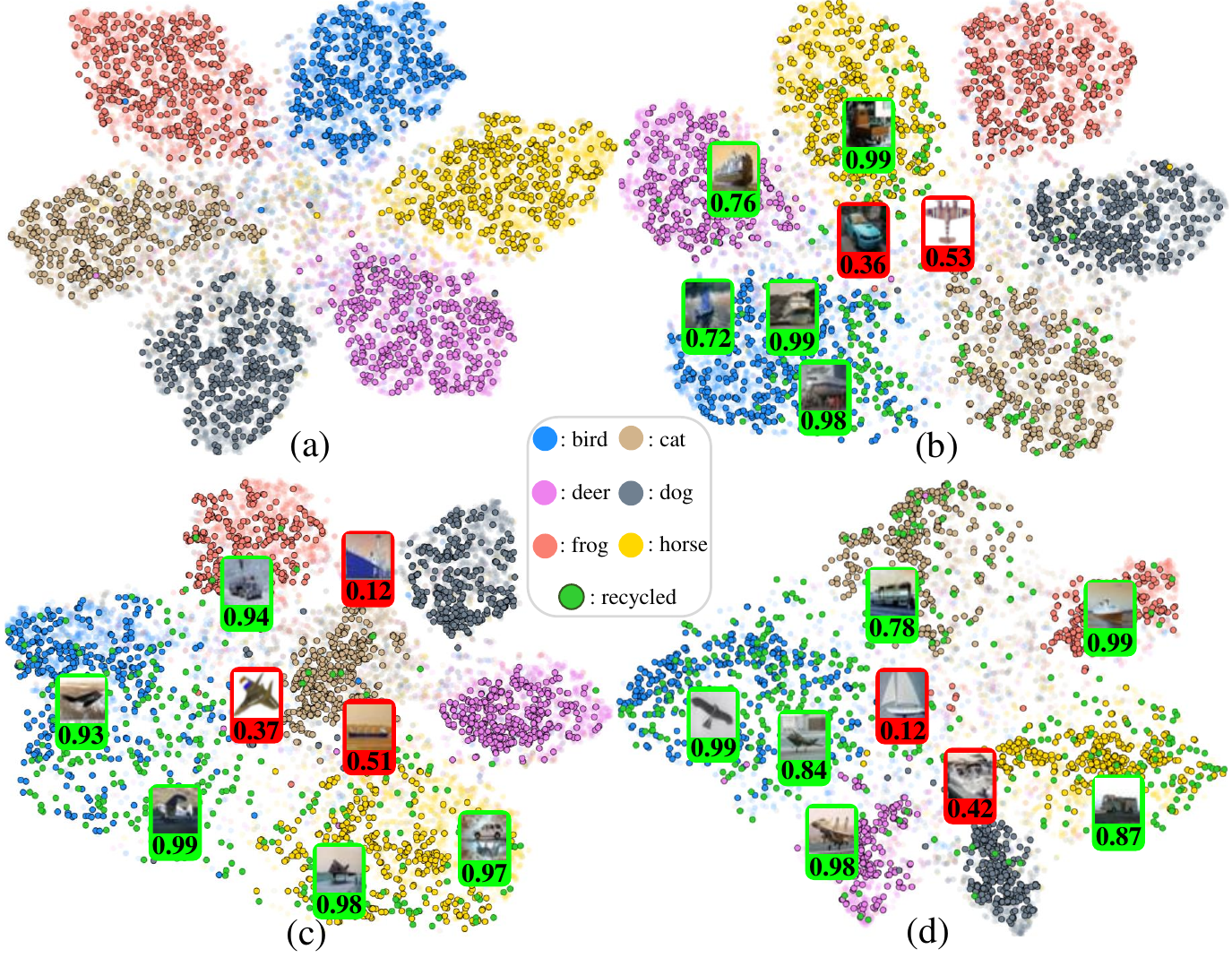}
	\end{center}
	\caption{t-SNE visualizations of the image representations. Figures (a)$\sim$(d) correspond to the results obtained on CIFAR10 dataset with $\zeta$ equaling 0\%, 25\%, 50\%, 75\%, respectively. Different colors denote the image data from different classes, where the green dots indicate the recycled OOD data determined by TOOR, and the dots with black boundary denote the originally labeled data. Besides, some of the recyclable OOD images (green boxes) and non-recyclable OOD images (red boxes) are visualized, of which the transferability scores $w(\mathbf{x})$ are also indicated below the images.}
	\label{fig:tsne_sup}
\end{figure}

Furthermore, we provide the visualization of the image features from SVHN and CIFAR10 datasets extracted by Wide ResNet-28-2 via using t-SNE method~\cite{maaten2008visualizing} in Figures~\ref{fig:tsne} and~\ref{fig:tsne_sup}, respectively. Here the result on CIFAR100+ImageNet dataset is not displayed as this dataset contains too many classes which is difficult to conduct performance visualization.

For SVHN dataset, it can be observed that most of the recycled OOD data (see the green dots) with relatively large transferability scores lie in the dense region within each cluster, while those with small transferability scores are distributed in a scattered way. Besides, we can see that the recycled OOD image data show great similarity to the classes in the labeled set. For example, we can see that many OOD data corresponding to digit ``9'' are mapped to the ID data with class ``0'', as these two digits look similar. In contrast, we can see that most of the non-recyclable OOD data in the red boxes are blurry and some of them cannot be recognized even by human. Hence, our method managed to alleviate the negative influence of these non-recyclable OOD data by assigning them with small transferability scores.

As for CIFAR10 dataset, we can see that most of the recycled OOD data (see the green dots) with relatively large transferability scores also lie in the dense region within each cluster, while those with small transferability scores are distributed in a scattered way. This finding is consistent with that on the above SVHN dataset. Moreover, the recycled OOD images also show great similarity to the classes in the labeled set. For instance, some of the ``airplane'' images are recycled to the ``bird'' classes, as they have similar shapes. On the other hand, some of the images that show less similarity to the animal classes are assigned with small transferability scores, such as ``0.12'' and ``0.42''. In a word, our strategy of utilizing transferability to characterize recyclability is also applicable to CIFAR10 dataset and helps to exploit the potential information inherited by OOD data in the mismatched classes.

\begin{table*}[t]
	\centering
	\setlength{\tabcolsep}{4.0mm}
	\caption{Ablation study on EMA smoothing and temperature scaling for OOD data detection.}
	\label{OOD_ablation_table}
	\begin{threeparttable}
		\begin{tabular}{ccccc}
			\toprule[1pt]
			&w/o EMA and temperature& 	with temperature&	with EMA&	TOOR   \\
			\midrule[0.6pt]
			SVHN& $88.97\pm0.16$&	$89.15\pm0.22$&	$91.10\pm0.18$&	$91.69\pm0.14$ \\
			CIFAR10& $76.49\pm0.20$&	$76.63\pm0.42$&	$77.97\pm0.52$&	$78.25\pm0.62$ \\
			CIFAR100+ImageNet& $45.28\pm0.65$&	$46.29\pm1.44$&	$47.95\pm0.87$&	$49.15\pm0.76$ \\
			\bottomrule[1pt]
		\end{tabular}
	\end{threeparttable}
\vspace{-2mm}
\end{table*}

\begin{table*}[t]
	\centering
	\setlength{\tabcolsep}{3.8mm}
	\caption{Ablation study on domain similarity score and class tendency score for OOD data Recycling.}
	\label{recycle_ablation_table}
	\begin{threeparttable}
		\begin{tabular}{cccccc}
			\toprule[1pt]
			&w/o recycle&	w/o two scores&	with domain similarity score&	with class tendency score&	TOOR   \\
			\midrule[0.6pt]
			SVHN& $89.08\pm0.19$&	$86.40\pm0.23$&	$89.52\pm0.25$&	$90.87\pm0.16$ &	$91.69\pm0.14$ \\
			CIFAR10& $75.74\pm0.23$&	$71.82\pm0.65$&	$76.37\pm0.54$&	$77.22\pm0.45$&	$78.25\pm0.62$ \\
			CIFAR100+ImageNet&$45.34\pm0.82$	&$43.16\pm1.19$	&$46.79\pm1.04$	&$48.20\pm0.93$	&$49.15\pm0.76$ \\
			\bottomrule[1pt]
		\end{tabular}
	\end{threeparttable}
\vspace{-2mm}
\end{table*}

\vspace{-2mm}
\subsection{Ablation Study}\label{ablation_study}
From the performance study in Section~\ref{performance}, we have shown the effectiveness of our OOD data detection and OOD data recycling. To further understand the contribution of our method, we conduct ablation studies to examine each module of the proposed OOD data detection and OOD data recycling. Here we keep all experimental settings as the same in the previous performance study.

First, our OOD data detection leverages EMA smoothing to stabilize the softmax score with temperature scaling. Here we decompose the stabilized softmax score of our OOD data detection into four experimental settings, namely ``w/o EMA and temperature'', ``with temperature'', ``with EMA'', and ``TOOR''. The experimental results over three independent runs on various datasets are shown in Table~\ref{OOD_ablation_table}. We can see that the softmax score with temperature only improves the performance marginally, but our EMA smoothing can make huge contribution to the results when compared with temperature scaling. Overall, our TOOR method combining EMA with temperature can achieve the best results.

Then, we conduct ablation study on the proposed domain similarity score and class tendency score to analyze the transferability evaluation during OOD data recycling. Specifically, we decompose the transferability of TOOR into four experimental settings, namely ``w/o two scores'', ``with domain similarity score'',	``with class tendency score'', and ``TOOR''. Here in the ``w/o two scores'' setting, we set the transferability scores of all OOD data to 1. Moreover, we add a compared baseline setting ``w/o recycle'' to denote training without recycling. The experimental results over three independent runs on various datasets are shown in Table~\ref{recycle_ablation_table}. We can see that without the transferability evaluated by either of the scores, the recycle process would hurt the learning performance. However, both scores can improve the performance when compared with the training without recycling. Moreover, the class tendency score brings more benefits to the experimental result than the domain similarity score. Furthermore, the combination of two scores enables our TOOR method to achieve the best results.

\section{Conclusion}
In this paper, we propose a novel SSL method termed TOOR that solves the class mismatch problem. Concretely, we utilize EMA to generate a stabilized softmax score to better detect the OOD data. Then, instead of discarding or down-weighting all the detected OOD data, we propose a novel weighting mechanism that integrates both domain information and label prediction knowledge to adaptively quantify the transferability of each OOD datum to find the transferable subset, which is further recycled via adversarial domain adaptation. As a result, the recycled OOD data can be re-used to help to train an improved semi-supervised classifier. The comparison results with various state-of-the-art SSL methods on various benchmark datasets firmly demonstrate the effectiveness of our proposed TOOR in handling class mismatch problems. In the future, we plan to study a more complex case when there exist some private classes in labeled data. In this case, the private classes owned by labeled data could mislead the learning of unlabeled data. Moreover, the private classes only contain scarce labeled data when compared with other classes with abundant unlabeled data, thus causing a class imbalance problem. Therefore, more advanced strategies for recyclable data identification should be developed to tackle such problem.

\bibliographystyle{IEEEtran}
\bibliography{bibfile}

\end{document}